\documentclass[10pt,twocolumn,letterpaper]{article}

\usepackage[pagenumbers]{cvpr} 

\usepackage{amssymb}
\usepackage[dvipsnames]{xcolor}
\usepackage{tabularx}
\usepackage{multirow} 
\usepackage{makecell} 
\usepackage{colortbl} 
\usepackage[Export]{adjustbox} 
\usepackage{arydshln} 
\usepackage{colortbl}
\usepackage{pifont} 
\usepackage{tikz} 
\usepackage{bm}
\usepackage{caption}
\usepackage{subcaption} 
\def\eg{\emph{e.g.}\xspace} 
\def\ie{\emph{i.e.}\xspace} 

\def\wrt{\emph{w.r.t.}\xspace}
\def\etal{\emph{et al.}\xspace}

\newcommand{\xmark}{\ding{55}}%
\newcommand{\noo}{\textcolor{red}{\xmark}}
\newcommand{\yes}{\textcolor{OliveGreen}{\checkmark}}

\newcommand{\ours}{LoopSplat\xspace}
\newcommand{\boldparagraph}[1]{\vspace{0.1em}\noindent{\bf #1}}

\colorlet{colorFst}{Green!25}       
\colorlet{colorSnd}{SpringGreen!45} 
\colorlet{colorTrd}{Yellow!30}      
\colorlet{colorLow}{darkgray!30}    
\newcommand{\fs}{\cellcolor{colorFst}\bf}   
\newcommand{\nd}{\cellcolor{colorSnd}}      
\newcommand{\rd}{\cellcolor{colorTrd}}      

\newcommand{\meanGS}[0]{\boldsymbol{\mu}}
\newcommand{\cov}[0]{\boldsymbol{\Sigma}}
\newcommand{\loss}{\mathcal{L}}

\newcolumntype{R}{>{\raggedleft\arraybackslash}X}

\usepackage{tocloft}

\definecolor{cvprblue}{rgb}{0.21,0.49,0.74}
\usepackage[pagebackref,breaklinks,colorlinks,citecolor=cvprblue]{hyperref}

\title{LoopSplat: Loop Closure by Registering 3D Gaussian Splats}

\author{Liyuan~Zhu\textsuperscript{1} \quad
Yue~Li\textsuperscript{2} \quad
Erik~Sandström\textsuperscript{3} \quad
Shengyu~Huang\textsuperscript{3} \quad
Konrad~Schindler\textsuperscript{3} \quad
Iro Armeni\textsuperscript{1} 
\quad
\vspace{5px}\\
{ \textsuperscript{1}{Stanford University} \quad 
\textsuperscript{2}{University of Amsterdam} \quad
\textsuperscript{3}{ETH Zurich} 
} \\
}

\begin{document}

\twocolumn[{%
	\renewcommand\twocolumn[1][]{#1}%
        \maketitle
	\begin{center}
    \begin{tabular}{ccccc}
    \multicolumn{5}{c}{%
        \includegraphics[width=\linewidth]{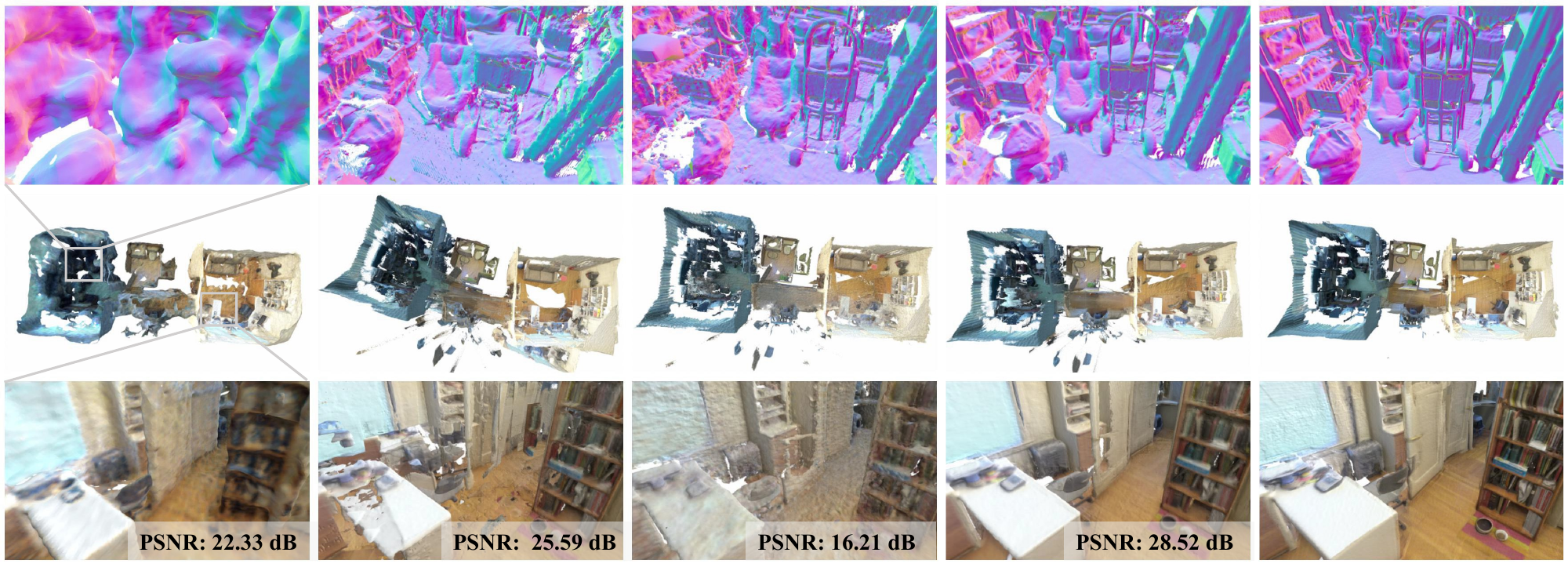}%
    }
     \\
     \hspace{1cm} \small GO-SLAM~\cite{zhang2023goslam} & \hspace{0.45cm} \small Gaussian-SLAM~\cite{yugay2023gaussian} & \hspace{0.4cm}\small Loopy-SLAM~\cite{liso2024loopyslam} & \hspace{0.7cm}\small  LoopSplat (Ours) & \small Ground Truth \\
\end{tabular}
	\captionof{figure}{\textbf{Dense Reconstruction on ScanNet~\cite{dai2017scannet}} \texttt{scene0054}. \ours demonstrates superior performance in geometric accuracy, robust tracking, and high-quality re-rendering. This is enabled by our globally consistent reconstruction approach utilizing 3DGS~\cite{kerbl20233d}.
 }
	\label{fig:teaser}
	\vspace{2mm}
	\end{center}    
}]
\maketitle
\begin{abstract}
Simultaneous Localization and Mapping (SLAM) based on 3D Gaussian Splats (3DGS) has recently shown promise towards more accurate, dense 3D scene maps. However, existing 3DGS-based methods fail to address the global consistency of the scene via loop closure and/or global bundle adjustment. To this end, we propose LoopSplat, which takes RGB-D images as input and performs dense mapping with 3DGS submaps and frame-to-model tracking. LoopSplat triggers loop closure online and computes relative loop edge constraints between submaps directly via 3DGS registration, leading to improvements in efficiency and accuracy over traditional global-to-local point cloud registration. It uses a robust pose graph optimization formulation and rigidly aligns the submaps to achieve global consistency. Evaluation on the synthetic Replica and real-world TUM-RGBD, ScanNet, and ScanNet++ datasets demonstrates competitive or superior tracking, mapping, and rendering compared to existing methods for dense RGB-D SLAM. Code is available at \href{https://loopsplat.github.io/}{loopsplat.github.io}.
\end{abstract}    
\section{Introduction}
\label{sec:intro}
Dense Simultaneous Localization and Mapping (SLAM) with RGB-D cameras has seen steady progress throughout the years from traditional approaches ~\cite{newcombe2011kinectfusion,niessner2013voxel_hashing,whelan2015elasticfusion,dai2017bundlefusion,schops2019bad,cao2018real} to neural implicit methods ~\cite{Sucar2021IMAP:Real-Time,zhu2022nice,sandstrom2023point,liso2024loopyslam,sandstrom2023uncle,zhang2023goslam,yang2022vox,eslam_cvpr23,wang2023co,mao2023ngel,tang2023mips} and recent methods that employ 3D Gaussians~\cite{kerbl20233d} as the scene representation~\cite{yugay2023gaussian,keetha2023splatam,matsuki2023monogs,yan2024gs,huang2024photo}. 
Existing methods can be split into two categories, \textit{decoupled} and \textit{coupled}, where \textit{decoupled} methods~\cite{huang2024photo,zhang2023goslam,chung2022orbeez,Rosinol2022NeRF-SLAM:Fields,matsuki2023newton} do not leverage the dense map for the tracking task, while the \textit{coupled} methods~\cite{yugay2023gaussian,keetha2023splatam,matsuki2023monogs,yan2024gs,Sucar2021IMAP:Real-Time,zhu2022nice,yang2022vox,mahdi2022eslam,wang2023co,sandstrom2023point,sandstrom2023uncle,tang2023mips,liso2024loopyslam} employ frame-to-model tracking using the dense map. Decoupling mapping and tracking generally creates undesirable redundancies in the system, such as inefficient information sharing and increased computational overhead. On the other hand, all \textit{coupled} 3DGS SLAM methods lack strategies for achieving global consistency on the map and the poses, which leads to an accumulation of pose errors and distorted maps.
Among the recent methods that enforce global consistency via loop closure and/or global bundle adjustment (BA), GO-SLAM~\cite{zhang2023goslam} requires costly retraining of the hash grid features to deform the map and Photo-SLAM~\cite{huang2024photo} similarly requires additional optimization of the 3D Gaussian parameters to resolve pose updates from the ORB-SLAM~\cite{Mur-Artal2017ORB-SLAM2:Cameras} tracker. These re-integration techniques need to save all mapped frames in memory, which limits their scalability. To avoid saving all mapped frames, Loopy-SLAM~\cite{liso2024loopyslam} uses submaps of neural point clouds and rigidly updates them after loop closure. However, to compute the loop edge constraints, Loopy-SLAM uses traditional global-to-local point cloud registration. This is not only slow, but also fails to leverage the property of the scene representation itself.

To address limitations of current systems, we seek a \textit{coupled} SLAM system that avoids saving all mapped input frames and is able to extract loop constraints directly from the dense map, without redundant compute. Framed as a research question, we ask: \textit{Can we use the map representation (i.e., 3DGS) itself for loop closure in a SLAM system?}
To this end, we propose a dense RGB-D SLAM system that uses submaps of 3D Gaussians for local frame-to-model tracking and dense mapping and is based on existing systems~\cite{yugay2023gaussian,matsuki2023monogs}. Different to the latter, we achieve global consistency via online loop closure detection and pose graph optimization. Importantly, we show that traditional point cloud registration techniques are not suitable to derive the loop edge constraints from 3D Gaussians and propose a new registration method that directly operates on the 3DGS representation, hence using 3DGS as a unified scene representation for tracking, mapping, and maintaining global consistency. Our key \textbf{contributions} are:

\begin{enumerate}
    \item We introduce \textit{\ours}, a coupled RGB-D SLAM system based on Gaussian Splatting, featuring a novel loop closure module. This module operates directly on Gaussian splats, integrating both 3D geometry and visual scene content for robust loop detection and closure.
    
    \item We develop an effective way to register two 3DGS representations, so as to efficiently extract edge constraints for pose graph optimization. Leveraging the fast rasterization of 3DGS, it is seamlessly integrated into the system, outperforming traditional techniques in terms of both speed and accuracy.

    \item We enhance the tracking and reconstruction performance of 3DGS-based RGB-D SLAM system, demonstrating marked improvements and increased robustness across diverse real-world datasets.
\end{enumerate}

\section{Related Work}
\label{sec:related_work}

\boldparagraph{Dense Visual SLAM.} 
The seminal work of Curless and Levoy~\cite{curless1996volumetric} paved the way for dense 3D mapping with truncated signed distance functions. Using frame-to-model tracking, KinectFusion~\cite{newcombe2011kinectfusion} showed that real-time SLAM is possible from a commodity depth sensor. To address the cubic memory scaling to the scene size, numerous works utilized voxel hashing ~\cite{niessner2013voxel_hashing,Kahler2015infiniTAM,Oleynikova2017voxblox,dai2017bundlefusion,matsuki2023newton} and octrees~\cite{steinbrucker2013large,yang2022vox,marniok2017efficient,chen2013scalable,liu2020neural} for map compression. Point-based representations have also been popular~\cite{whelan2015elasticfusion,schops2019bad,cao2018real,Kahler2015infiniTAM,keller2013real,cho2021sp,zhang2020dense,sandstrom2023point,liso2024loopyslam,zhang2024glorie}, with surfels and lately using neural points or 3D Gaussians~\cite{yugay2023gaussian,keetha2023splatam,yan2024gs,matsuki2023monogs,zhang2024glorie,sandstrom2024splat}. 
To tackle the issue of accumulating pose errors, globally consistent dense SLAM methods have been developed, where a subdivision of the global map into submaps is common~\cite{cao2018real,dai2017bundlefusion,fioraio2015large,tang2023mips,matsuki2023newton,maier2017efficient,kahler2016real,stuckler2014multi,choi2015robust,Kahler2015infiniTAM,reijgwart2019voxgraph,henry2013patch,bosse2003atlas,maier2014submap,tang2023mips,mao2023ngel,liso2024loopyslam}, followed by pose graph optimization~\cite{cao2018real,maier2017efficient,tang2023mips,matsuki2023newton,kahler2016real,endres2012evaluation,engel2014lsd,kerl2013dense,choi2015robust,henry2012rgb,yan2017dense,schops2019bad,reijgwart2019voxgraph,henry2013patch,stuckler2014multi,wang2016online,matsuki2023newton,hu2023cp,mao2023ngel,liso2024loopyslam} to deform the submaps between them. Additionally, some works employ global BA for refinement~\cite{dai2017bundlefusion,schops2019bad,cao2018real,teed2021droid,yan2017dense,yang2022fd,matsuki2023newton,chung2022orbeez,tang2023mips,hu2023cp,zhang2023goslam}. 3D Gaussian SLAM with RGB-D input has also been shown, however, methods fail to consider global consistency~\cite{yugay2023gaussian,keetha2023splatam,yan2024gs,matsuki2023monogs}, leading to error accumulation in the map and pose estimates. Most similar to our work is Loopy-SLAM~\cite{liso2024loopyslam}, which uses the explicit neural point cloud representation of Point-SLAM~\cite{sandstrom2023point} and equips it with global consistency via loop closure on submaps. \ours differentiates itself from Loopy-SLAM  and demonstrates improvements in three key areas: \textit{(i)} We improve the accuracy and efficiency of the relative pose constraints by directly registering 3DGS, instead of resorting to classical techniques like FPFH~\cite{rusu2009fpfh} with RANSAC, followed by ICP~\cite{besl1992method}. \textit{(ii)} We avoid having to mesh the submaps in a separate process for registration and use the 3D Gaussians directly. \textit{(iii)} For loop detection, we rely on a combination of image matching and overlap between submaps, leading to better detections than using only image content as in \cite{liso2024loopyslam}.

\begin{figure*}[t]
    \centering
    \includegraphics[width=\linewidth]{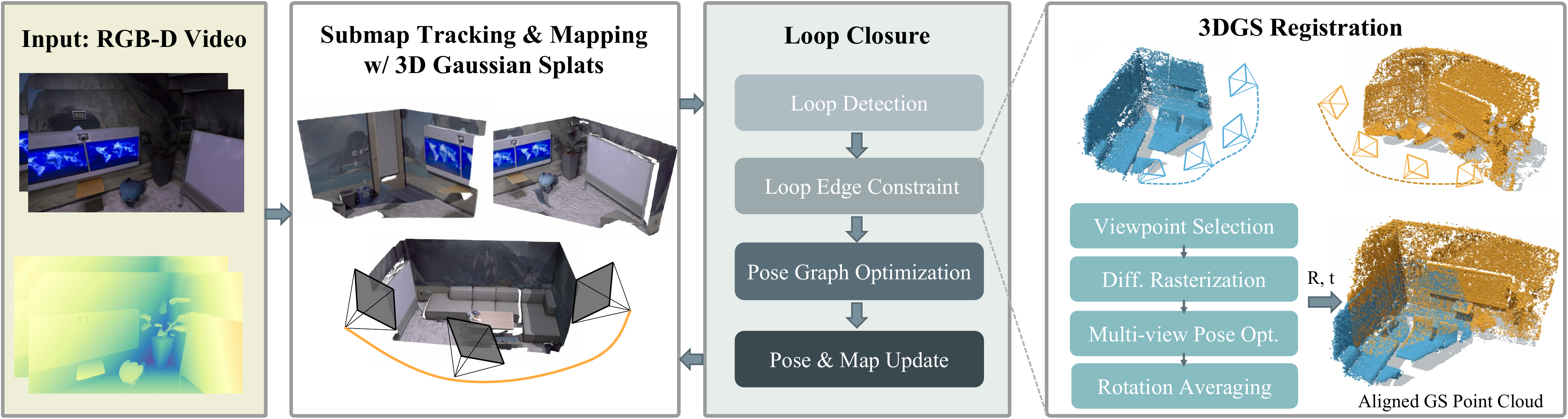}
\vspace{-5px}
    \caption{\textbf{\ours Overview.} LoopSplat is a \textit{coupled} RGB-D SLAM system that uses Gaussian splats as a \textbf{unified} scene representation for tracking, mapping, and maintaining global consistency. In the front-end, it continuously estimates the camera position while constructing the scene using Gaussian splats. When the camera traverses beyond a predefined threshold, the current submap is finalized, and a new one is initiated. Concurrently, the back-end loop closure module monitors for location revisits. Upon detecting a loop, the system generates a pose graph, incorporating loop edge constraints derived from our proposed \textbf{3DGS registration}. Subsequently, pose graph optimization (PGO) is executed to refine both camera poses and submaps, ensuring overall spatial coherence.
    }
    \label{fig:overview}
    \vspace{-3px}
\end{figure*}
\boldparagraph{Geometric Registration.} 
Geometric registration is an important component of building edge constraints for pose graphs. Specifically, point cloud registration aims to find a rigid transformation that aligns two point cloud fragments into the same coordinate framework. Traditional methods leverage hand-crafted local descriptors~\cite{rusu2009fpfh,tombari2010unique} for feature matching, followed by RANSAC for pose estimation. Recent learning-based methods either use patch-based local descriptors~\cite{zeng20173dmatch,gojcic2019perfect} or efficient fully-convolutional ones~\cite{Choy2019FCGF,bai2020d3feat}. BUFFER~\cite{ao2023buffer} balances the efficiency and generalization of local descriptors by combining fully-convolutional backbones for key-point detection with a patch-based network for feature description. To address fragment registration with low overlap, Predator~\cite{huang2021predator} uses attention mechanisms~\cite{vaswani2017attention} to guide key-point sampling, significantly improving the robustness of algorithms. This has been further enhanced through coarse-to-fine matching~\cite{qin2023geotransformer}. Point clouds lack the continuous, view-dependent, and multi-scale representation capabilities of NeRFs, limiting their ability to fully capture complex 3D scene in SLAM.

Neural Radiance Fields (NeRF)~\cite{mildenhall2021nerf} have been widely adopted for various applications beyond scene reconstruction, including scene understanding~\cite{engelmann2024opennerf}, autonomous driving~\cite{wang2023fegr}, and SLAM~\cite{sucar2021imap,pan2024pin_slam}. When modeling large-scale scenes with NeRF, it is necessary to partition the scene into blocks to manage memory constraints and to ensure sufficient representation power. Consequently, registering NeRFs to merge different partitions emerged as a research problem. 
iNeRF~\cite{yen2020inerf} aligns a query image to the NeRF map through analysis-by-synthesis: it optimizes the camera pose so that the rendered image matches the query. However, this method is only suitable for local refinement due to its non-convex nature, which can cause the model to get stuck in local minima. NeRF2NeRF~\cite{goli2023nerf2nerf} aims to align two NeRFs by extracting surface points from the density field and aligning manually selected keypoints to estimate the pose. DReg-NeRF~\cite{chen2023dreg} addresses NeRF registration similarly to point cloud registration, by first extracting surface points and then applying a fully convolutional feature extraction backbone. Recently, Gaussian Splatting~\cite{kerbl20233d} has started to replace NeRFs due to its efficient rasterization and flexible editing capabilities, afforded by the explicit representation. GaussReg \cite{chang2024gaussreg} pioneered learning-based 3D Gaussian Splatting (3DGS) registration, drawing on the fast rendering of 3DGS. However, all previous NeRF and 3DGS registration methods \cite{yen2020inerf,goli2023nerf2nerf,chen2023dreg,chang2024gaussreg} assume ground truth camera poses for training views, which is not suitable for real-world SLAM scenarios. Moreover, these methods have only explored pairwise registration in small-scale scenes. Our method, without any training or preprocessing, directly operates on estimated camera poses from the SLAM front-end and can be integrated into loop closure on the fly.

\section{\ours}
\label{sec:method}
\ours is an RGB-D SLAM system that simultaneously estimates the camera poses and builds a 3D Gaussian map from input frames in a globally consistent manner. This section begins with a recap of the Gaussian-SLAM system described in \cite{yugay2023gaussian} (\cref{sec:front-end}) -- which is the base of \ours, followed by the introduction of the proposed 3DGS registration module (\cref{sec:registration}). Finally, the integration of loop closure into the Gaussian-SLAM system, enabled by the registration module, is presented in~\cref{sec:loopclose}. Please see~\cref{fig:overview} for an overview of the proposed system. 

\subsection{Gaussian Splatting SLAM}
\label{sec:front-end}
We follow \cite{yugay2023gaussian, liso2024loopyslam} and represent the scene using a collection of submaps, each modeling several keyframes with a 3D Gaussian point cloud $\mathbf{P}^s$, where
\begin{equation}
\mathbf{P}^s = \{G_{i}(\mu, \Sigma, o, C)| , i=1,\ldots,N\},
\end{equation}
with individual Gaussian mean $\mu \in \mathbb{R}^3$, covariance matrix $\Sigma \in \mathbb{R}^{3 \times 3}$, opacity value $o \in \mathbb{R}$, and RGB color $C \in \mathbb{R}^3$.

\paragraph{Submap Initialization.} 
Starting from the first keyframe $\mathbf{I}^s_f$, each submap models a sequence of keyframes observing a specific region. As the explored scene space expands, a new submap is initialized to avoid processing the entire global map simultaneously. Unlike previous approaches that use a fixed number of keyframes \cite{choi2015robust, dai2017bundlefusion, maier2014submap}, we dynamically trigger new submap initialization when the current frame's relative displacement or rotation to the first keyframe $\mathbf{I}^s_f$ exceeds the predefined thresholds, $d_\text{thre}$ or $\theta_\text{thre}$.

\paragraph{Frame-to-model Tracking.}
To localize a incoming frame $\mathbf{I}^s_j$ within the current submap $\mathbf{P}^s$, we first initialize the camera pose $\mathbf{T}_j$ based on the constant motion assumption as: $\mathbf{T}_{j} = \mathbf{T}_{j-1} \cdot \mathbf{T}_{j-2}^{-1} \cdot \mathbf{T}_{j-1}$. Next, we optimize $\mathbf{T}_j$ by minimizing the tracking loss $\loss_{\text{tracking}}(\hat{\mathbf{I}}_j^s, \hat{\mathbf{D}}^s_j, \mathbf{I}_j^s, \mathbf{D}_j^s, \mathbf{T}_j)$, which measures the discrepancy between the rendered color $\hat{\mathbf{I}}_j$ and depth $\hat{\mathbf{D}}^s_j$ images at viewpoint $\mathbf{T}_j$, and the input color $\mathbf{I}_j^s$ and depth $\mathbf{D}_j^s$.
To stabilize tracking, we use an alpha mask $M_\text{a}$ and an inlier mask $M_\text{in}$ to address gross errors caused by poorly reconstructed or previously unobserved areas. The final tracking loss is a sum over the valid pixels as
\begin{equation}
\loss_\text{tracking} = \sum M_\text{in} \cdot M_\text{a} \cdot(\lambda_{c}|\hat{\mathbf{I}}_j^s - \mathbf{I}_j^s|_1 + (1 - \lambda_{c})|\hat{\mathbf{D}}_j^s - \mathbf{D}_j^s|_1),
\end{equation}
where $\lambda_c$ is a weight that balances the color and depth losses, and $\| \cdot\|$ denotes the $\loss_1$ loss between two images. Please refer to the supplementary material for more details.

\paragraph{Submap Expansion.} Keyframes are selected by fixed interval for the submap. Once the current keyframe $\mathbf{I}_j^s$ is localized, we expand the 3D Gaussian map primarily in sparsely covered regions for efficient mapping. We first compute a posed dense point cloud from the RGB-D input and then uniformly sample $M_k$ points from areas where the accumulated alpha values are below a threshold $\alpha_{\text{thre}}$ or where significant depth discrepancies occur. These points are initialized as anisotropic 3D Gaussians, with scales defined based on the nearest neighbor distance within the current submap. New 3D Gaussian splats are added to the current submap only if there is no existing 3D Gaussian mean within a radius $\rho$.

\paragraph{Submap Update.} \label{sec:submap_mapping}
After new Gaussians are added, all Gaussians in the current submap are optimized for a fixed number of iterations by minimizing the rendering loss $\loss_{\text{render}}$, computed over all keyframes of the submap, with at least 40\% of the compute allocated to the most recent keyframe.
The rendering loss is of three components: color loss $\loss_{\text{color}}$, depth loss $\loss_{\text{depth}}$, and a regularization term $\loss_{\text{reg}}$:
\begin{equation}
\loss_{\text{render}} = \lambda_\text{color} \cdot \loss_\text{color} + \lambda_\text{depth} \cdot \loss_\text{depth} + \lambda_\text{reg} \cdot \loss_\text{reg},
\label{eq}
\end{equation}
where $\lambda_*$ are hyperparamters.
Similar to the tracking loss, the depth loss is the $\loss_1$ loss between rendered and ground truth depth maps. For color supervision, we use a weighted combination of the $\loss_1$ and $\mathrm{SSIM}$~\cite{wang2004image} loss:
\begin{equation}
\loss_\text{col} = (1 - \lambda_{SSIM}) \cdot |\hat{\mathbf{I}} - \mathbf{I}|_1 +
\lambda_{SSIM} \big(1 - \mathrm{SSIM}(\hat{\mathbf{I}}, \mathbf{I}) \big),
\label{eq
}
\end{equation}
where $\lambda_{SSIM} \in [0,1]$.
To regularize overly elongated 3D Gaussians in sparsely covered or barely observed regions, we add an isotropic regularization term~\cite{matsuki2023monogs}
\begin{equation}
\loss_\text{reg} = \frac{1}{K}\sum_{k \in K}|s_k - \overline{s}_k|_1,
\end{equation}
where $s_k\in\mathbb{R}^3$ is the scale of a 3D Gaussian, $\overline{s}_k$ is its mean, and $K$ is the number of Gaussians in the submap.
During optimization, to preserve geometry directly measured from the depth sensor and reduce computation time, we do not clone or prune the Gaussians~\cite{kerbl20233d}.

\subsection{Registration of Gaussian Splats}
\label{sec:registration}
LoopSplat's first contribution relates to the registration of Gaussian splats which is formulated as following. Consider two overlapping 3D Gaussian submaps $\mathbf{P}$ and $\mathbf{Q}$, each reconstructed using different keyframes and not aligned. The goal is to estimate a rigid transformation $\mathbf{T}_{\mathbf{P}\rightarrow \mathbf{Q}} \in SE(3)$ that aligns $\mathbf{P}$ with $\mathbf{Q}$. Each submap is also associated with a set of viewpoints $\mathbf{V^P}$ as: 
\begin{equation}
\mathbf{V^P} = \{\mathbf{v}^\mathbf{p}_i=(\mathbf{I}, \mathbf{D}, \mathbf{T})_i|i=0,\ldots, N\},
\end{equation}
where $\mathbf{I}$ and $\mathbf{D}$ are the individual RGB and depth measurements, respectively, and $\mathbf{T}$ is the estimated camera pose in~\cref{sec:front-end}.

\paragraph{Overlap Estimation.} \label{para:overlap}
Knowing the approximate overlap between the source and target submaps $\mathbf{P}$ and $\mathbf{Q}$ is crucial for robust and accurate registration, and this co-contextual information can be extracted by comparing feature similarities~\cite{huang2021predator}. While the means of the Gaussian splats do form a point cloud, we found that estimating the overlap region directly from them by matching local features does not work well (\cf \cref{sec:ablation}).
Instead, we identify viewpoints from each submap that share similar visual content. Specifically, we first pass all keyframes through NetVLAD~\cite{arandjelovic2016netvlad} to extract their global descriptors. We then compute the cosine similarity between the two sets of keyframes and retain the top-$k$ pairs for registration.

\paragraph{Registration as Keyframe Localization.} 
Given that the 3DGS submap and its viewpoints can be treated as one rigid body, we propose to approach 3DGS registration as a keyframe localization problem. For a selected viewpoint $\mathbf{v}^\mathbf{p}_i$, determining its camera pose $\mathbf{T}_i^{q}$ within $\mathbf{Q}$ allows one to render the same RGB-D image from $\mathbf{Q}$ as $\mathbf{v}^\mathbf{p}_i$. Hence, the rigid transformation $\mathbf{T}_{\mathbf{P}\rightarrow \mathbf{Q}}$ can be computed as $\mathbf{T}_i^q \cdot \mathbf{T}_i^{-1}$. 

During keyframe localization, we keep the parameters of $\mathbf{Q}$ fixed and optimize the rigid transformation $\mathbf{T}_{\mathbf{P}\rightarrow \mathbf{Q}}$ by minimizing the rendering loss $\loss = \loss_{col} + \loss_{depth}$~\cite{matsuki2023gaussian},
where both $\loss_{col}$ and $\loss_{depth}$ are $\loss_1$ losses.

We estimate the rigid transformations for the selected viewpoints, from $\mathbf{P}$ to $\mathbf{Q}$ for viewpoints in $\mathbf{V^P}$ and vice versa for $\mathbf{V^Q}$, in parallel. The rendering residuals $\epsilon$ are also saved upon completion of the optimization. By using the sampled top-$k$  viewpoints from the estimated overlap region as the selected viewpoints, the registration efficiency is greatly improved without redundancy in non-overlapping viewpoints.
Viewpoint transformations are estimated first, then used to compute the submap's global transformation.

\paragraph{Multi-view Pose Refinement.} Given a set of transformations $\{(\mathbf{T}_{\mathbf{P} \rightleftharpoons \mathbf{Q}},\varepsilon )_i\}_{i=1}^{2k}$,  where the first $k$ estimates are from $\mathbf{P}\rightarrow\mathbf{Q}$ and the last $k$ estimates from $\mathbf{Q}\rightarrow\mathbf{P}$, one must find a global consensus for the transformation $\bar{\mathbf{T}}_{\mathbf{P}\rightarrow \mathbf{Q}}$. As the rendering residual indicates how well the transformed viewpoint fits the original observation, we take the reciprocal of the residuals as a weight for each estimate and apply weighted rotation averaging~\cite{peretroukhin2020rotation_averaging,bregier2021roma} to compute the global rotation:
\begin{equation}
    \bar{\mathbf{R}} = \arg \min_{\mathbf{R} \in SO3} \sum_{i=1}^{k} \frac{1}{\varepsilon_i}\|\mathbf{R} - \mathbf{R}_i\|_F^2 + \sum_{i=k+1}^{2k} \frac{1}{\varepsilon_i}\|\mathbf{R} - \mathbf{R}_i^{-1}\|_F^2,
\end{equation}
where $\| \cdot\|_F^2$ denotes the Frobenius norm. The global translation is found as the weighted mean over individual estimates.

\subsection{Loop Closure with 3DGS}
\label{sec:loopclose}
Loop closure aims to identify pose corrections (\ie relative transformations \wrt the current estimates) for past submaps and keyframes to ensure global consistency. This process is initiated when a new submap is created, and upon detecting a new loop, the pose graph, which includes all historical submaps, is constructed. The loop edge constraints for the pose graph are then computed using 3DGS registration (\cref{sec:registration}). Subsequently, Pose Graph Optimization (PGO)~\cite{choi2015robust} is performed to achieve globally consistent multi-way registration of 3DGS.

\paragraph{Loop Closure Detection.} To effectively detect system revisits to the same place, we first extract a global descriptor $\mathbf{d}\in \mathbb{R}^{1024}$ using a pretrained NetVLAD~\cite{arandjelovic2016netvlad}. We compute the cosine similarities of all keyframes within the $i$-th submap and determine the self-similarity score $s_\mathrm{self}^i$ corresponding to their $\mathrm{p}$-th percentile. We then apply the same method to compute the cross-similarity $s_\mathrm{cross}^{i,j}$ between the $i$-th and $j$-th submaps. A new loop is added if $s_\mathrm{cross}^{i,j} > \mathrm{min}(s_\mathrm{self}^i, s_\mathrm{self}^j)$.
However, relying solely on visual similarity for loop closure~\cite{liso2024loopyslam} can generate false loop edges, potentially degrading PGO performance. To mitigate that risk, we additionally evaluate the initial geometric overlap ratio $r$~\cite{huang2021predator} between the Gaussians of two submaps, and retain only loops with $r>0.2$. See Supp. for more details. 

\paragraph{Pose Graph Optimization.}
We create a new pose graph every time a new loop is detected and ensure that its connections match the previous one, besides the new edges introduced by the new submap. The relative pose corrections $\{\mathbf{T}_{c^i} \in SE(3)\}$ to each submap are defined as nodes in the pose graph, which are connected with odometry edges and loop edges. Here $\mathbf{T}_{c^i}$ denotes the correction applied to $i$-th submap. The nodes and edges connecting adjacent nodes (\ie, odometry edges) are initialized with identity matrices. Loop edge constraints are added at detected loops and initialized according to the Gaussian splatting registration (\cref{sec:registration}). The information matrices for edges are computed directly from the Gaussian centers and incorporated into the pose graph. PGO is triggered after loop detection and we use a robust formulation based on line processes~\cite{choi2015robust}.

\paragraph{Globally Consistent Map Adjustment.} From the PGO output, we obtain a set of pose corrections $\{\mathbf{T}_{c^i} = [\mathbf{R}_{c^i}| \mathbf{t}_{c^i}]\}_{i=1}^{N_s}$ for $N_s$ submaps, with $c_i$ denoting correction for submap $i$. For each submap, we update camera poses, the Gaussian means and covariances
\begin{gather}
    \mathbf{T}_j \leftarrow \mathbf{T}_{c^i} \mathbf{T}_j \enspace, \\
    \meanGS_i  \leftarrow \mathbf{R}_{c^i} \meanGS_{\mathbf{S}^i} + \mathbf{t}_{c^i}, \enspace
    \cov_i  \leftarrow \mathbf{R}_{c^i} \cov_{\mathbf{S}^i} \mathbf{R}_{c^i}^T \enspace.
\end{gather}
Here, $\meanGS_i$ and $\cov_i$ represent the sets of centers and covariance matrices, respectively, of the Gaussians in the $i$-th submap $\mathbf{S}^i$, index $j$ is iterated over the keyframe span of the submap. We omit spherical harmonics (SH) to reduce the Gaussian map size and improve pose estimation accuracy~\cite{matsuki2023monogs}.
\section{Experiments}
\begin{table}[t]
\centering
\setlength{\tabcolsep}{2pt}
\renewcommand{\arraystretch}{1.05}
\resizebox{\columnwidth}{!}
{
\begin{tabular}{lcccccccccc}
\toprule
Method & LC & \texttt{Rm 0} & \texttt{Rm 1} & \texttt{Rm 2} & \texttt{Off 0} & \texttt{Off 1} & \texttt{Off 2} & \texttt{Off 3} & \texttt{Off 4} & Avg.\\
\midrule
\multicolumn{11}{l}{\cellcolor[HTML]{EEEEEE}{\textit{Neural Implicit Fields}}} \\ 
\multirow{1}{*}{NICE-SLAM~\cite{zhu2022nice}} & \noo &  0.97
& 1.31 &  1.07  &  0.88 & 1.00  &  1.06  &  1.10  & 1.13 &  1.06 \\[0.8pt] \noalign{\vskip 1pt}

\multirow{1}{*}{Vox-Fusion~\cite{yang2022vox}} & \noo &1.37 & 4.70 &  1.47 & 8.48 & 2.04  &  2.58 & 1.11 & 2.94 & 3.09\\[0.8pt]  \noalign{\vskip 1pt}
\multirow{1}{*}{ESLAM~\cite{eslam_cvpr23}} & \noo &0.71 & 0.70 & 0.52  & 0.57  & 0.55  & 0.58  & 0.72 &  0.63 & 0.63 \\[0.8pt]  \noalign{\vskip 1pt}
\multirow{1}{*}{Point-SLAM~\cite{sandstrom2023point}
} 
& \noo & 0.61  &  0.41  & 0.37   &  0.38 &  0.48 & 0.54  &  0.69 &  0.72 &  0.52 \\
[0.8pt]  \noalign{\vskip 1pt}
\multirow{1}{*}{MIPS-Fusion~\cite{mipsfusion_siga23} 
} 
& \yes &1.10  &  1.20  & 1.10   &  0.70  & 0.80   & 1.30  & 2.20 &  1.10 & 1.19  \\
[0.8pt]  \noalign{\vskip 1pt}

\multirow{1}{*}{GO-SLAM~\cite{zhang2023goslam} 
} 
& \yes &  0.34  &  0.29 &   0.29   & \rd  0.32  &  0.30  &  \rd 0.39 & 0.39 & \rd 0.46& 0.35 \\
[0.8pt]  \noalign{\vskip 1pt}
\multirow{1}{*}{Loopy-SLAM~\cite{liso2024loopyslam}} 
& \yes &\fs 0.24  & \rd 0.24  & \rd 0.28 & \nd 0.26  &  0.40  &  \nd 0.29 & \rd 0.22  & \nd 0.35  & \nd 0.29  \\
[0.8pt]

\hdashline \noalign{\vskip 1pt}
\multicolumn{11}{l}{\cellcolor[HTML]{EEEEEE}{\textit{3D Gaussian Splatting}}} \\ 
\multirow{1}{*}{SplaTAM~\cite{keetha2023splatam} 
} 
& \noo &  0.31  &  0.40   &  0.29   &  0.47   &  0.27   &  \nd 0.29  & 0.32  &  0.72 &  0.38  \\
[0.8pt]  \noalign{\vskip 1pt}
\multirow{1}{*}{MonoGS~\cite{matsuki2023monogs} 
} 
& \noo & 0.33  &  \fs 0.22  & 0.29    & 0.36    &  \nd 0.19  & \fs 0.25  & \fs 0.12 & 0.81  & 0.32  \\
[0.8pt]  \noalign{\vskip 1pt}
\multirow{1}{*}{Gaussian-SLAM~\cite{yugay2023gaussian} 
} 
& \noo & \rd 0.29  &  0.29  & \nd 0.22    & 0.37    &  \rd 0.23  &  0.41  &  0.30 & \nd 0.35  &\rd  0.31  \\
[0.8pt]  \noalign{\vskip 1pt}
\multirow{1}{*}{{$^*$}Photo-SLAM~\cite{huang2024photo} 
} 
& \yes &  0.54  &  0.39   &  0.31   &  0.52   &  0.44   &   1.28  & 0.78  &  0.58 &  0.60  \\
[0.8pt]  \noalign{\vskip 1pt}
\multirow{1}{*}{\textbf{\ours (Ours)}} 
& \yes &\nd 0.28  & \fs 0.22  & \fs 0.17 & \fs 0.22  & \fs 0.16  &   0.49 & \nd 0.20  & \fs 0.30  & \fs 0.26  \\

\bottomrule
\end{tabular}
}
\caption{\textbf{Tracking Performance on Replica~\cite{replica19arxiv}} (ATE RMSE $\downarrow$ [cm]). LC indicates loop closure. The best results are highlighted as \colorbox{colorFst}{\bf first}, \colorbox{colorSnd}{second}, and \colorbox{colorTrd}{third}. \ours performs the best. $^*$Photo-SLAM~\cite{huang2024photo} is a \textit{decoupled} method using ORB-SLAM3~\cite{campos2021orb_slam3} for tracking and loop closure.}
\label{tab:replica_tracking}
\end{table}

\begin{table}[t]
\centering
\scriptsize
\begin{tabularx}{\columnwidth}{lXXXXXc}
\toprule
\textbf{Method} & \texttt{a} & \texttt{b} & \texttt{c} & \texttt{d} & \texttt{e}  & \textbf{Avg.} \\
\midrule
\multicolumn{7}{l}{\cellcolor[HTML]{EEEEEE}{\textit{Neural Implicit Fields}}} \\ 
Point-SLAM~\cite{sandstrom2023point} & 246.16 & 632.99 & 830.79 & 271.42 & 574.86 & 511.24 \\
ESLAM~\cite{eslam_cvpr23} & 25.15 & \nd 2.15 & 27.02 & 20.89 & 35.47 & 22.14 \\
GO-SLAM~\cite{zhang2023goslam} & 176.28 & 145.45 & 38.74 & 85.48	&106.47	& 110.49 \\
Loopy-SLAM~\cite{eslam_cvpr23} & N/A & N/A & 25.16 & 234.25 & 81.48 & 113.63 \\ [0.8pt]
\hdashline \noalign{\vskip 1pt}
\multicolumn{7}{l}{\cellcolor[HTML]{EEEEEE}{\textit{3D Gaussian Splatting}}} \\
SplaTAM~\cite{keetha2023splatam} & \rd 1.50 & \fs 0.57 & \fs 0.31 & 443.10 & \rd 1.58 & 89.41 \\
MonoGS~\cite{yugay2023gaussian} & 7.00 & 3.66 & \rd 6.37 & \nd3.28 & 44.09 & \rd 12.88 \\
Gaussian SLAM~\cite{yugay2023gaussian} &\nd 1.37 &\rd 2.82 & 6.80 & \rd 3.51 & \fs 0.88 & \nd 3.08 \\
\textbf{\ours (Ours)} & \fs 1.14 & 3.16 & \nd 3.16 & \fs1.68 & \nd0.91 & \fs2.05 \\
\bottomrule
\end{tabularx}
\caption{\textbf{Tracking Performance on ScanNet++~\cite{yeshwanthliu2023scannetpp}} (ATE RMSE $\downarrow$ [cm]). \ours achieves the highest accuracy and can robustly deal with the large camera motions in the sequence.}
\label{tab:scannetpp_tracking}
\end{table}

\begin{table}[t]
\centering
\setlength{\tabcolsep}{5pt}
\renewcommand{\arraystretch}{1.1}
\resizebox{\columnwidth}{!}
{
\begin{tabular}{llllllllllll}
\toprule

Method & \texttt{00} & \texttt{59} & \texttt{106} & \texttt{169} & \texttt{181} & \texttt{207} &  \texttt{54} & \texttt{233} &   Avg. \\ \midrule
\multicolumn{11}{l}{\cellcolor[HTML]{EEEEEE}{\textit{Neural Implicit Fields}}} \\ 
Vox-Fusion~\cite{yang2022vox} & 16.6 & 24.2 & 8.4 & 27.3 &  23.3 & 9.4 & -  & - & -\\
 Co-SLAM~\cite{wang2023co} & 7.1  & 11.1   & 9.4  & \fs 5.9 & 11.8 & 7.1 &  -  & - &- \\
MIPS-Fusion~\cite{mipsfusion_siga23} & 7.9  & 10.7   & 9.7  & 9.7 & 14.2 & 7.8  & - &- & -\\
NICE-SLAM~\cite{zhu2022nice} &12.0 &  14.0 &  7.9 &10.9 & 13.4 &\nd 6.2  &  20.9 &9.0  & 13.0\\
 ESLAM~\cite{eslam_cvpr23} & 7.3   &  8.5   & \rd 7.5  & \nd 6.5 & \rd 9.0 & \fs 5.7  & 36.3 & \fs 4.3 & 10.6  \\ 
 Point-SLAM~\cite{sandstrom2023point} &  10.2   &  7.8    & 8.7  & 22.2 &  14.8 &  9.5  & 28.0 & 6.1  & 14.3 \\ 
 GO-SLAM~\cite{zhang2023goslam} & \nd 5.4  & \nd 7.5   & \fs 7.0  & 7.7 & \fs 6.8 &  6.9  & \nd 8.8 & \rd 4.8 & \fs 6.9 \\ 
Loopy-SLAM~\cite{liso2024loopyslam} & \fs 4.2   & \nd 7.5   & 8.3  & \rd 7.5 &  10.6 & 7.9  & \fs 7.5 &  5.2  & \nd 7.7 \\ [0.8pt]

\hdashline \noalign{\vskip 1pt}
\multicolumn{11}{l}{\cellcolor[HTML]{EEEEEE}{\textit{3D Gaussian Splatting}}} \\ 
MonoGS~\cite{matsuki2023monogs} & 9.8& 32.1 & 8.9 & 10.7 & 21.8 & 7.9   & 17.5 & 12.4  & 15.2 \\
SplaTAM~\cite{keetha2023splatam} & 12.8  & 10.1  & 17.7  & 12.1 &  11.1 & 7.5  & 56.8 & \rd 4.8  & 16.6 \\
Gaussian-SLAM~\cite{yugay2023gaussian} & 21.2& 12.8   & 13.5  & 16.3 & 21.0 & 14.3  & 37.1  & 11.1 & 18.4 \\
\textbf{\ours (Ours)} & \rd 6.2   & \fs 7.1  & \nd 7.4  & 10.6 & \nd 8.5 & \rd6.6  & \rd 16.0 & \nd 4.7  & \rd 8.4 \\
\bottomrule 
\end{tabular}
}
\caption{\textbf{Tracking Performance on ScanNet~\cite{dai2017scannet}}. \ours outperforms 3DGS-based systems by a large margin and is on par with the state-of-the-art baselines.}
\label{tab:scannet_tracking}
\end{table}
\label{sec:results}
Here we describe our experimental setup and compare our method to state-of-the-art baselines. We evaluate tracking, reconstruction, and rendering performance on synthetic and real-world datasets, with a dedicated ablation study for loop closure. For implementation details, please refer to Supp.

\paragraph{Datasets.} 
We evaluate on four datasets: \emph{Replica}~\cite{straub2019replica} is a synthetic dataset with high-quality 3D indoor reconstructions. We use the same RGB-D sequences as~\cite{Sucar2021IMAP:Real-Time}. \emph{ScanNet}~\cite{dai2017scannet} is a real-world dataset with its poses estimated by BundleFusion~\cite{dai2017bundlefusion}. We evaluate on eight scenes with loops following~\cite{liso2024loopyslam,zhang2023goslam}. \emph{ScanNet++}~\cite{yeshwanthliu2023scannetpp} is a real, high-quality dataset. We use five DSLR-captured sequences where poses are estimated with COLMAP~\cite{schoenberger2016colmap} and refined with the help of laser scans. \emph{TUM-RGBD}~\cite{Sturm2012ASystems} is a real-world dataset with accurate poses obtained from a motion capture system.

\paragraph{Baselines.} We compare \ours with state-of-the-art \textit{coupled} RGB-D SLAM methods, categorized into two groups based on the underlying scene representation:
\emph{(i)} Neural implicit fields: MIPS-Fusion~\cite{mipsfusion_siga23}, GO-SLAM~\cite{zhang2023goslam}, and Loopy-SLAM~\cite{liso2024loopyslam}, all of which incorporate loop closure; and 
\emph{(ii)} 3DGS:  MonoGS~\cite{matsuki2023monogs}, SplaTAM~\cite{keetha2023splatam}, Gaussian-SLAM~\cite{yugay2023gaussian}, and Photo-SLAM~\cite{huang2024photo}.
For completeness, we include Photo-SLAM~\cite{huang2024photo} in our evaluation, noting that it utilizes ORB-SLAM3~\cite{campos2021orb_slam3} for tracking and loop closure, setting it apart from all other tested methods. 

\begin{table}[t]
\centering
\setlength{\tabcolsep}{6pt}
\renewcommand{\arraystretch}{1.05}
\resizebox{\columnwidth}{!}
{
\begin{tabular}{lcllllll}
\toprule
   \multirow{2}{*}{Method} & \multirow{2}{*}{LC} & \texttt{fr1/} &  \texttt{fr1/} & \texttt{fr1/} & \texttt{fr2/} & \texttt{fr3/} & \multirow{2}{*}{Avg.} \\
  & & \texttt{desk} &  \texttt{desk2} & \texttt{room} & \texttt{xyz} & \texttt{off.} &  \\ 
\midrule
\multicolumn{8}{l}{\cellcolor[HTML]{EEEEEE}{\textit{Neural Implicit Fields}}} \\ 
DI-Fusion~\cite{huang_difusion}         & \noo & 4.4  &  N/A & N/A & 2.0 & 5.8 &  N/A\\
NICE-SLAM~\cite{zhu2022nice}         & \noo & 4.26 &  4.99  &34.49  &6.19  &3.87 &10.76 \\
Vox-Fusion~\cite{yang2022vox}        & \noo & 3.52 & 6.00  &  19.53 & 1.49 &  26.01 &   11.31 \\ 
MIPS-Fusion~\cite{mipsfusion_siga23}      & \yes & 3.0  & N/A  & N/A & 1.4 &  4.6 &  N/A \\ 
Point-SLAM~\cite{sandstrom2023point} & \noo & 4.34 & 4.54  &  30.92  &  1.31  &3.48 &  8.92 \\
ESLAM~\cite{eslam_cvpr23}          & \noo & 2.47 &  3.69  & 29.73 &  1.11 &  2.42 &  7.89 \\ 
Co-SLAM~\cite{wang2023co}            & \noo & 2.40 & N/A  & N/A & 1.70 &  2.40 &  N/A \\ 
GO-SLAM~\cite{zhang2023goslam}       & \yes & \fs 1.50  & N/A  & \fs 4.64 & \rd 0.60 &  \rd 1.30 &  N/A \\ 
Loopy-SLAM~\cite{liso2024loopyslam}      & \yes & 3.79  & \nd 3.38  &  7.03 & 1.62 & 3.41 & \rd 3.85  \\

\hdashline \noalign{\vskip 1pt}
\multicolumn{8}{l}{\cellcolor[HTML]{EEEEEE}{\textit{3D Gaussian Splatting}}} \\
SplaTAM~\cite{keetha2023splatam}     & \noo & 3.35  & 6.54  &  11.13 &  1.24 &  5.16 &  5.48\\
MonoGS~\cite{matsuki2023monogs}      & \noo & \nd 1.59  & 7.03  & 8.55 & 1.44 &  1.49 &  4.02 \\
Gaussian-SLAM~\cite{yugay2023gaussian}      & \noo & 2.73  & 6.03  & 14.92 & 1.39 &  5.31 &  6.08 \\
{$^*$}Photo-SLAM~\cite{huang2024photo}     & \yes & 2.60  & N/A  & N/A & \fs 0.35 &  \nd1.00 &  N/A \\
\textbf{\ours (Ours)}             & \yes &  2.08  & \rd 3.54  &   6.24 & 1.58 & 3.22 & \nd 3.33  \\[0.8pt]
%
\hdashline \noalign{\vskip 1pt}
\multicolumn{8}{l}{\cellcolor[HTML]{EEEEEE}{\textit{Classical}}} \\
BAD-SLAM~\cite{schops2019bad}       & \yes & 1.7 &  N/A &   N/A &  1.1 & 1.7 &  N/A\\
Kintinuous~\cite{whelan2012kintinuous} & \yes & 3.7  & 7.1 &   7.5  & 2.9 & 3.0 &   4.84\\
ORB-SLAM2~\cite{Mur-Artal2017ORB-SLAM2:Cameras} & \yes & \rd 1.6 &   \fs 2.2 & \nd 4.7 & \nd 0.4 & \nd 1.0 &  \fs 1.98\\
ElasticFusion~\cite{whelan2015elasticfusion} & \yes &  2.53 & 6.83 & 21.49 &  1.17 & 2.52 & 6.91\\
BundleFusion~\cite{dai2017bundlefusion} & \yes & \rd 1.6 &  N/A &   N/A & 1.1 & 2.2 &  N/A\\
Cao~\etal~\cite{cao2018real}        & \yes & \fs 1.5 &  N/A &   N/A & \rd 0.6 & \fs 0.9 &  N/A\\
Yan~\etal~\cite{yan2017dense}       & \yes & \rd 1.6 &  N/A &  \rd 5.1 &  N/A & 3.1 &  N/A\\
%
\bottomrule
\end{tabular}
}
\caption{\textbf{Tracking Performance on TUM-RGBD~\cite{sturm2012benchmark}} (ATE RMSE $\downarrow$ [cm]). 
$^*$ indicates using ORB-SLAM3~\cite{campos2021orb_slam3} for tracking and loop closure. \ours performs the best among \textit{coupled} SLAM, further closing the gap to sparse solver-based SLAM.}
\label{tab:tum_tracking}
\end{table}

\paragraph{Evaluation Metrics.}
\emph{Tracking} accuracy is measured by the root mean square absolute trajectory error (ATE RMSE)~\cite{Sturm2012ASystems}.
For \emph{reconstruction}, we follow \cite{sandstrom2023point} and evaluate via meshes extracted with marching cubes~\cite{lorensen1987marching}, using a voxel size of $1$ cm. We measure rendered mesh depth error at sampled novel views as in~\cite{zhu2022nice} and the F1-score, \ie, the harmonic mean of precision and recall \wrt ground truth mesh vertices.
\emph{Rendering} quality is evaluated by comparing full-resolution rendered images to input training views in terms of PSNR, SSIM~\cite{wang2004image}, and LPIPS~\cite{zhang2018unreasonable}. We note that comparing to training views may yield too optimistic results, but it enables a consistent comparison with existing methods.
To assess \emph{map size}, we measure the total memory needed for the map and the peak GPU memory usage.
\emph{Runtime} is reported as average per-frame tracking and map optimization time, as well as loop edge registration runtime.

\subsection{Tracking}
\begin{figure}[t]
    \centering

    \begin{tabular}{ccc}
    \multicolumn{3}{c}{%
        \includegraphics[width=\linewidth]{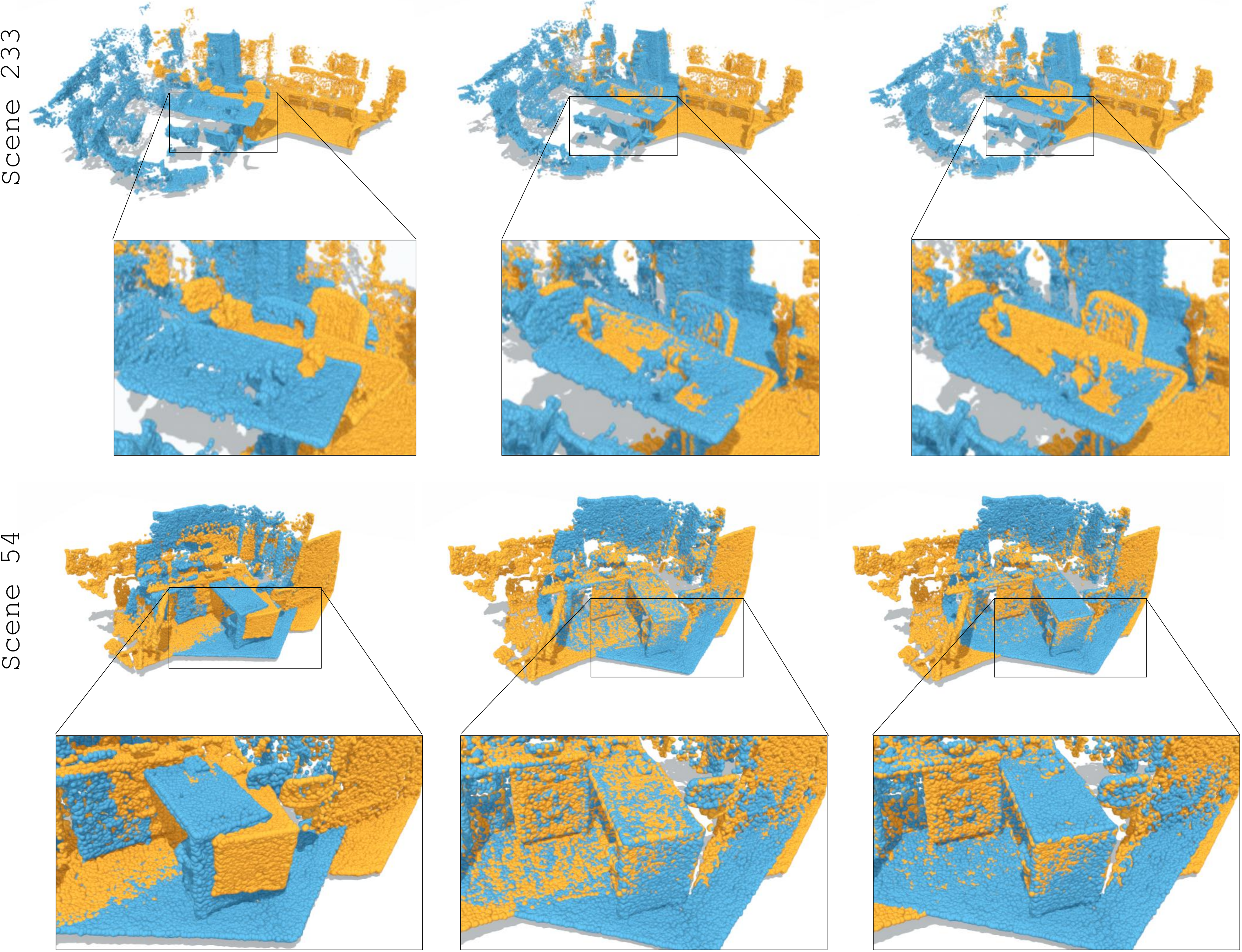}%
    } \\
    \small{\hspace{0.2cm}Gaussian-SLAM~\cite{yugay2023gaussian}} & \small{\hspace{0.5cm} Ours} & \small{\hspace{0.7cm}Ground Truth} \\
\end{tabular}

    \caption{\textbf{Comparison of Submap Alignment on ScanNet~\cite{dai2017scannet}}. We visualize the centers of 3D Gaussians as point clouds. Two submaps are colorized differently. \ours consistently aligns the submaps better than Gaussian-SLAM~\cite{yugay2023gaussian}. 
    }
    \label{fig:reg_lc}
\end{figure}

We report the camera tracking performance in \cref{tab:replica_tracking,tab:scannet_tracking,tab:scannetpp_tracking,tab:tum_tracking}. On Replica, we outperform all the baselines, achieving a 10\% higher accuracy compared to the second best one. On real-world datasets, we achieve the highest pose accuracy on TUM-RGBD and ScanNet++ among all neural implicit field-based and 3DGS-based baselines, improving tracking accuracy by 14\% and 33\%, respectively. 
It is worth noting that, for all 3DGS-based baselines~\cite{matsuki2023monogs,yugay2023gaussian,keetha2023splatam}, trajectory errors accumulate as trajectories grow longer in larger scenes with loops and motion blur, \eg, ScanNet \texttt{00}, \texttt{59}, \texttt{181} and TUM-RGBD \texttt{fr1/desk2} and \texttt{fr1/room}. We attribute our superior tracking performance to the robust 3DGS registration that underpins our loop closure. On ScanNet, we obtain the third-best performance. We note that the ground truth poses in ScanNet, derived from BundleFusion~\cite{dai2017bundlefusion}, appear to have limited accuracy: visual inspection suggests that our method achieves better alignment and reconstruction than the ground truth; see \cref{fig:teaser}, \cref{fig:reg_lc}, and \texttt{scene 233} in \cref{fig:recon}. Additional qualitative examples are in Supp. Besides superior tracking accuracy, our coupled method avoids redundant computations for separate tracking and map reconstruction, in contrast to \textit{decoupled} ones like GO-SLAM~\cite{zhang2023goslam} and Photo-SLAM~\cite{huang2024photo}.

\subsection{Reconstruction}
\begin{figure*}[t]
    \centering
 \begin{tabular}{ccccc}
    \multicolumn{5}{c}{%
        \includegraphics[width=\linewidth]{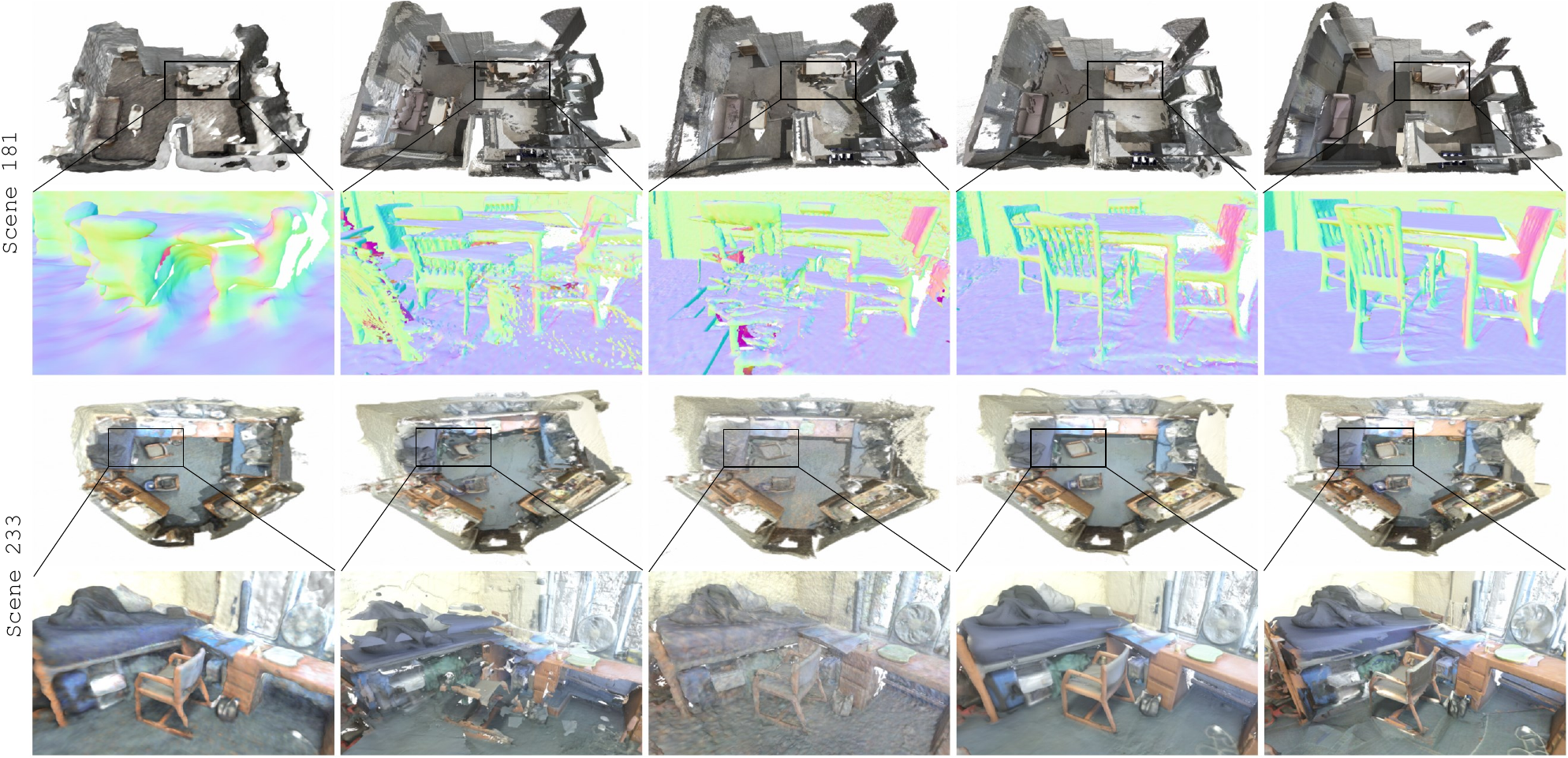}%
    } \\
    \hspace{1cm} \small GO-SLAM~\cite{zhang2023goslam} & \hspace{0.45cm} \small Gaussian-SLAM~\cite{yugay2023gaussian} & \hspace{0.4cm}\small Loopy-SLAM~\cite{liso2024loopyslam} & \hspace{0.7cm}\small  LoopSplat (Ours) & \small Ground Truth \\
\end{tabular}  
\vspace{-2mm}
    \caption{\textbf{Comparison of Mesh Reconstruction on two ScanNet~\cite{dai2017scannet} scenes}. For the first scene, we highlight shape details with normal shading, showing that \ours yields the best geometry (\eg the chairs). For the second one, we display renderings of the colored mesh. Note the distortions at the desk in ground truth that are not present in ours, indicating accuracy limitations of ScanNet ground truth poses.
    }
    \label{fig:recon}
\label{fig:scannet_recon}

\end{figure*}

We evaluate the mesh reconstruction quality on Replica, the only dataset with high accuracy ground truth mesh, in \cref{tab:replica_recon}\footnote{{$^*$}Depth L1 for GO-SLAM is based on results reproduced by \cite{liso2024loopyslam} using random poses, as GO-SLAM originally evaluates on ground truth poses.}. \ours outperforms all 3DGS-based baselines attributed to more accurate pose estimates. LoopSplat falls behind Loopy-SLAM~\cite{liso2024loopyslam} and Point-SLAM~\cite{sandstrom2023point}, but note that the latter two require ground truth depth to determine where to sample points during ray-marching, thus assuming perfect input depth. 
\cref{fig:scannet_recon} compares ScanNet meshes reconstructed with \ours to those of the best-performing baselines, GO-SLAM and Loopy-SLAM (both also including loop closure), as well as to a 3DGS baseline, Gaussian-SLAM (which does not perform loop closure). Our method recovers more geometric details (\eg, on the chairs). On ScanNet \texttt{233}, the visual quality and completeness of our reconstruction appears even better than the ground truth, especially on the floor, desk and bed.

\begin{table}[t]

\centering
\setlength{\tabcolsep}{2pt}
\renewcommand{\arraystretch}{1.2}
\resizebox{\columnwidth}{!}
{
\begin{tabular}{llccccccccc}
\toprule
Method & Metric & \texttt{Rm\thinspace0} & \texttt{Rm\thinspace1} & \texttt{Rm\thinspace2} & \texttt{Off\thinspace0} & \texttt{Off\thinspace1} & \texttt{Off\thinspace2} & \texttt{Off\thinspace3} & \texttt{Off\thinspace4} & Avg.\\
\midrule
\multicolumn{11}{l}{\cellcolor[HTML]{EEEEEE}{\textit{Neural Implicit Fields}}} \\ 
\multirow{2}{*}{\makecell[l]{NICE-\\SLAM~\cite{zhu2022nice}}}
& Depth L1 [cm] $\downarrow$ & 1.81  &  1.44	&  2.04	&  1.39	&  1.76	&8.33	&4.99	&2.01	&2.97 \\
& F1 [$\%$] $\uparrow$ & 45.0	&  44.8 & 43.6	& 50.0	& 51.9	& 39.2	& 39.9	& 36.5	& 43.9\\[0.8pt] \hdashline \noalign{\vskip 1pt}
\multirow{2}{*}{\makecell[l]{Vox-\\Fusion~\cite{yang2022vox}}} 
& Depth L1 [cm] $\downarrow$ & 1.09 & 1.90 & 2.21  & 2.32 & 3.40  &  4.19  & 2.96 & 1.61 & 2.46\\
& F1 [$\%$] $\uparrow$  & 69.9 &  34.4 &  59.7 &  46.5 &   40.8 &  51.0 & 64.6 & 50.7 &  52.2\\[0.8pt] \hdashline \noalign{\vskip 1pt}
\multirow{2}{*}{ESLAM~\cite{eslam_cvpr23}} & Depth L1 [cm] $\downarrow$ &  0.97  & 1.07  &  1.28  &  0.86 &  1.26  &  1.71  &  1.43  &  1.06 &  1.18 \\
& F1 [$\%$] $\uparrow$  & 81.0 &  82.2  &  83.9  &  78.4 &  75.5  &  77.1  &  75.5  &  79.1 &  79.1 \\ [0.8pt] \hdashline \noalign{\vskip 1pt}
\multirow{1}{*}{Co-SLAM~\cite{wang2023co}} & Depth L1 [cm] $\downarrow$ & 1.05 & 0.85  & 2.37  & 1.24 & 1.48  & 1.86  & 1.66  & 1.54 & 1.51 \\[0.8pt] \hdashline \noalign{\vskip 1pt}
\multirow{3}{*}{\makecell[l]{GO-\\SLAM~\cite{zhang2023goslam}}} & Depth L1 [cm] $\downarrow$ & - & -  & -  & - & -  & -  & -  & - & 3.38 \\
& $^*$Depth L1 [cm] $\downarrow$ & 4.56 & 1.97  & 3.43  & 2.47 & 3.03  & 10.3  & 7.31  & 4.34 & 4.68 \\
& F1 [$\%$] $\uparrow$  & 17.3 & 33.4  & 24.0  & 43.0 & 31.8  & 21.8  & 17.3  & 22.0 & 26.3\\[0.8pt] \hdashline \noalign{\vskip 1pt}
\multirow{2}{*}{\makecell[l]{Point-\\SLAM~\cite{sandstrom2023point}}} 
& Depth L1 [cm] $\downarrow$ &  0.53 & 0.22 \nd  & 0.46 \nd    & \nd  0.30  &   0.57  &  \fs 0.49 & \nd 0.51 &  0.46 & \nd 0.44 \\
& F1 [$\%$] $\uparrow$  & 86.9   & 92.3 \nd  & \nd 90.8    & \nd  93.8  &  \fs  91.6  &  \fs   89.0 & \rd  88.2  &   85.6 & \rd  89.8 \\
[0.8pt] \hdashline \noalign{\vskip 1pt}
\multirow{2}{*}{Loopy-SLAM~\cite{liso2024loopyslam}} 
& Depth L1 [cm] $\downarrow$ &\fs 0.30& \fs 0.20& \fs  0.42 & \fs 0.23 &  \fs 0.46 &  \nd 0.60 & \fs 0.37& \fs 0.24& \fs 0.35\\
& F1 [$\%$] $\uparrow$  & \fs 91.6  & \fs 92.4 & \rd 90.6  & \fs  93.9 &  \fs 91.6  &  \rd 88.5  & \fs  89.0 & \fs 88.7 & \fs 90.8 \\
[0.8pt] 

\midrule

\multicolumn{11}{l}{\cellcolor[HTML]{EEEEEE}{\textit{3D Gaussian Splatting}}} \\ 

\multirow{2}{*}{\makecell[l]{SplaTAM~\cite{keetha2023splatam}}} 
& Depth L1 [cm]$\downarrow$ & \rd0.43 & 0.38 &  0.54  & 0.44 & 0.66 & 1.05 & 1.60 & 0.68  & 0.72 \\
& F1 [$\%$]$\uparrow$ & \rd 89.3 & 88.2 & 88.0 & 91.7 & 90.0 & 85.1 & 77.1 & 80.1  & 86.1 \\
\hdashline \noalign{\vskip 1pt}
\multirow{2}{*}{\makecell[l]{Gaussian SLAM~\cite{yugay2023gaussian} }} 
& Depth L1 [cm]$\downarrow$ & 0.61 &  0.25 &  0.54 & 0.50 & \rd 0.52 &  0.98 & 1.63 & \rd 0.42 &  0.68 \\
& F1 [$\%$]$\uparrow$  &  88.8 &  91.4 &  90.5 & 91.7 &  90.1 & 87.3 & 84.2 & \rd 87.4 & 88.9 \\
\hdashline \noalign{\vskip 1pt}

\multirow{2}{*}{\textbf{\ours (Ours)}} 
& Depth L1 [cm] $\downarrow$ & \nd 0.39 & \rd0.23& \rd 0.52 & \rd 0.32 & \nd 0.51  & \rd 0.63 & \rd 1.09& \nd 0.40 & \rd 0.51\\
& F1 [$\%$] $\uparrow$  & \nd 90.6  & \rd 91.9  & \fs 91.1   &  \rd 93.3  &  \rd 90.4   &  \nd 88.9   & \nd 88.7   &\nd  88.3  &  \nd 90.4  \\

\bottomrule
\end{tabular}
}
\label{fig:replica_recon}
\caption{\textbf{Reconstruction Performance on Replica~\cite{replica19arxiv}.} \ours obtains the second-best F1-score, falling behind only to Loopy-SLAM. It is noteworthy that both the NeRF-based Loopy-SLAM and Point-SLAM methods require ground truth depth input to guide the depth rendering, whereas our method, leveraging 3DGS, only requires estimated camera poses at rendering time. }

\label{fig:replica_recon_full}
\label{tab:replica_recon}
\vspace{1mm}
\end{table}

\begin{table}[t]
    \setlength{\tabcolsep}{2pt}
    \renewcommand{\arraystretch}{1.2}
	\centering
	\resizebox{\linewidth}{!}{
   \begin{tabular}{l|ccc|ccc|ccc}
    \toprule
    \textbf{Dataset} & \multicolumn{3}{c|}{\textbf{\emph{Replica}}~\cite{replica19arxiv}} & \multicolumn{3}{c|}{\textbf{\emph{TUM}}~\cite{sturm2012benchmark}} & \multicolumn{3}{c}{\textbf{\emph{ScanNet}}~\cite{dai2017scannet}} \\
     \midrule
      Method  &  PSNR $\uparrow$ &  SSIM $\uparrow$ &LPIPS $\downarrow$ &  PSNR $\uparrow$ &  SSIM $\uparrow$ &LPIPS $\downarrow$ &  PSNR $\uparrow$ &  SSIM $\uparrow$ &LPIPS $\downarrow$\\
    \midrule
   
    NICE-SLAM~\cite{zhu2022nice} &24.42& 0.892 & 0.233& 14.86& 0.614 & 0.441 & 17.54 & 0.621 & 0.548  \\
    Vox-Fusion~\cite{yang2022vox} &24.41 & 0.801 & 0.236 &16.46 & 0.677 &0.471 & 18.17 & 0.673 & 0.504 \\
   ESLAM~\cite{eslam_cvpr23} & 28.06 & 0.923 &0.245 & 15.26 &0.478 & 0.569&  15.29& 0.658 & \rd 0.488 \\
     Point-SLAM~\cite{sandstrom2023point}  & 35.17 &0.975  & 0.124& \rd 16.62 & \rd 0.696 & \rd 0.526 & \nd 19.82& \nd 0.751& 0.514\\
     Loopy-SLAM~\cite{liso2024loopyslam} & \nd 35.47 & \nd 0.981  & \nd 0.109 & 12.94 &  0.489 & 0.645 &15.23 & 0.629 & 0.671 \\
    SplaTAM~\cite{keetha2023splatam} & \rd 34.11 & \rd 0.970 &\fs 0.100 & \fs 22.80 & \fs 0.893& \fs 0.178& \rd 19.14 & \rd 0.716 & \fs 0.358 \\
    \textcolor{gray}{Gaussian-SLAM~\cite{yugay2023gaussian}} & \textcolor{gray}{42.08} & \textcolor{gray}{0.996} & \textcolor{gray}{0.018} & \textcolor{gray}{25.05}& \textcolor{gray}{0.929} & \textcolor{gray}{0.168}& \textcolor{gray}{27.67} & \textcolor{gray}{0.923} & \textcolor{gray}{0.248} \\
   \textbf{\ours (Ours)} & \fs 36.63 &\fs 0.985 & \rd 0.112& \nd 22.72 &\nd 0.873 &\nd 0.259 & \fs 24.92 &  \fs 0.845  & \nd 0.425\\
    \bottomrule
    \end{tabular}}
\caption{\textbf{Rendering Performance on 3 Datasets}. \ours achieves competitive results on synthetic and real-world datasets. \textcolor{gray}{Gray} indicates evaluation on submaps instead of a global map.}
\label{tab:rendering}
\vspace{-3mm}
\end{table}

\subsection{Rendering}
\cref{tab:rendering} reports our rendering performance on training views. To conduct a fair comparison, we merge all the submaps into a global one and optimize the global map with estimated cameras pose, to avoid local overfitting on submaps\footnote{Gaussian-SLAM evaluates rendering on local submaps.}. \ours surpasses all competing methods in terms of PSNR and LPIPS on Replica and ScanNet, and is competitive with SplaTAM on TUM-RGBD. Note the significant margin over baselines that employ implicit neural representations. We report the per-scene rendering results in Supp.

\subsection{Memory and Runtime Analysis}
\label{sec:exp-memory}


\cref{tab:memory_runtime} profiles the runtime and memory usage of \ours. While our per-frame tracking and map optimization time falls behind the fastest baselines, our Gaussian Splatting-based registration significantly shortens the loop edge registration time compared to Loopy-SLAM. Through careful control of submap growth, our Gaussian splats embedding size is 8$\times$ smaller than that of the 3DGS baseline SplaTAM. Additionally, we require the least GPU memory to process a room-sized scene. In contrast, baselines like ESLAM, GO-SLAM or SplaTAM require \textgreater15 GB of GPU memory.

\begin{table}[t]
  \centering
  \scriptsize
  \setlength{\tabcolsep}{4pt}
  \renewcommand{\arraystretch}{1.2}
  \resizebox{\columnwidth}{!}{
  \begin{tabular}{lccccc}
    \toprule
    Method  & Tracking & Mapping & Registration & Embedding & Peak GPU  \\
     & /Frame(s) $\downarrow$ & /Frame(s) $\downarrow$ & /Edge(s) $\downarrow$&  Size(MiB)$\downarrow$ & Use(GiB)$\downarrow$  \\
    \midrule
    NICE-SLAM~\cite{zhu2022nice}  & 1.06  & 1.15  & - &95.9 & 12.0  \\
    Vox-Fusion~\cite{yang2022vox} & 1.92  & 1.47  & -& \fs \textbf{0.15} & 17.6 \\
    Point-SLAM~\cite{sandstrom2023point} & 1.11  & 3.52 & - &   \nd27.2 & \nd 7.7 \\
    ESLAM~\cite{mahdi2022eslam} &  \nd0.15  & \nd0.62  &-&  \rd45.5  & 17.3  \\
    GO-SLAM~\cite{zhang2023goslam} &  \multicolumn{2}{c}{ \fs \textbf{0.125 }} & -& 48.1 & 18.4  \\
    SplaTAM~\cite{keetha2023splatam} & 2.70 & 4.89 &-& 404.5 & 18.5\\
    Loopy-SLAM\cite{liso2024loopyslam} & 1.11 & 3.52  & \nd 12.0  &   60.9 & \rd9.3 \\
    \textbf{\ours (Ours)} & \rd0.83   & \rd0.93  & \fs 1.36  & 49.7 & \fs \textbf{7.0} \\
    \bottomrule
  \end{tabular}
  }
  \caption{\textbf{Runtime and Memory Usage on Replica} \texttt{office 0}. Per-frame runtime is calculated as the total optimization time divided by the sequence length, profiled on a RTX A6000 GPU. The embedding size is the total memory of the map representation. Note that implicit field-based methods require additional space for their decoders. We take runtime values from \cite{yugay2023gaussian} and embedding values from \cite{liso2024loopyslam} for the baselines.}
  \label{tab:memory_runtime}
\vspace{-3mm}
\end{table}

\subsection{Ablations} \label{sec:ablation}
We first demonstrate that straightforward point cloud registration is not suitable to derive loop edge constraints from 3DGS. To illustrate this, we replace the proposed 3DGS registration in our SLAM system with FPFH+ICP~\cite{zhou2016fgr} and evaluate the trajectory error (ATE) on Replica. As shown in the last row of \cref{tab:ablation}, FPFH+ICP applied directly to the center points of 3D Gaussians leads to less accurate loop edges compared to our method and deteriorates loop closure. We hypothesize that this is because the center points do not accurately represent the scene surfaces, as previously discussed in \cite{Huang2DGS2024,Yu2024GOF,zhang2024rade}. 
Furthermore, the pre-processing of \cite{zhou2016fgr} involves re-rendering and back-projecting 3DGS to obtain 3D points, downsampling the point clouds and voxelizing them. This heavy pre-processing makes \cite{liso2024loopyslam} more than 8$\times$ slower than our method. In contrast, \ours efficiently reuses the native map representation without any pre-processing, answering the research question we asked in \cref{sec:intro}.
We also explore the impact of different modules in our registration method. The ablation study confirms that every component contributes to the final performance: Multi-view optimization and rotation averaging greatly improve registration accuracy by fusing information from different viewpoints.  View selection via overlap estimation (\cref{para:overlap}) is crucial to identify informative viewpoints and ensure the efficiency of the SLAM system.
\begin{table}[t]
    \setlength{\tabcolsep}{15pt}
    \renewcommand{\arraystretch}{1.05}
    \centering	
    \resizebox{\columnwidth}{!}{
    \begin{tabular}{ccccc}
    \toprule
    \textit{Mul. Opt.} & \textit{Ove. Est.} &  \textit{Rot. Ave.} & ATE (cm) & Runtime (s)  \\
    \midrule
    \noo & \noo & \noo & 0.31 &  - \\
    \noo & \yes & \noo & \rd 0.31 & \fs 1.25 \\
    \yes & \yes & \noo & \nd0.27 & \nd 1.36\\
    \yes & \noo & \yes & 0.37 & 11.02 \\
    \yes & \yes & \yes & \fs0.26 & \nd 1.36\\
    \multicolumn{3}{c}{\cellcolor{gray!25}FPFH+ICP~\cite{zhou2016fgr}} & 0.40 & 12.0 \\

    \bottomrule
    \end{tabular}
    } 
    \caption{\textbf{Ablation Study on 3DGS Registration}. The numbers are computed based on average performance of 8 scenes on Replica~\cite{straub2019replica}. \textit{Mul. Opt.} denotes multi-view optimization, \textit{Ove. Est.} and \textit{Rot. Ave.} denote view selection and rotation averaging.} 
    \label{tab:ablation}
\vspace{-3mm}
\end{table}

\section{Conclusion}
We presented \ours, a novel dense RGB-D SLAM system that exclusively uses 3D Gaussian Splats for scene representation, achieving global consistency through loop closure. Built around 3DGS submaps, \ours enables dense mapping, frame-to-model tracking, and online loop closure via direct 3DGS submap registration. Comprehensive evaluation on four datasets shows competitive or superior performance in tracking, mapping, and rendering. We discuss limitations and future work in Supp.

{
    \small
    \bibliographystyle{ieeenat_fullname}
    \bibliography{main}

\begin{thebibliography}{103}
\providecommand{\natexlab}[1]{#1}
\providecommand{\url}[1]{\texttt{#1}}
\expandafter\ifx\csname urlstyle\endcsname\relax
  \providecommand{\doi}[1]{doi: #1}\else
  \providecommand{\doi}{doi: \begingroup \urlstyle{rm}\Url}\fi

\bibitem[Ao et~al.(2023)Ao, Hu, Wang, Xu, and Guo]{ao2023buffer}
Sheng Ao, Qingyong Hu, Hanyun Wang, Kai Xu, and Yulan Guo.
\newblock Buffer: Balancing accuracy, efficiency, and generalizability in point cloud registration.
\newblock In \emph{CVPR}, 2023.

\bibitem[Arandjelovic et~al.(2016)Arandjelovic, Gronat, Torii, Pajdla, and Sivic]{arandjelovic2016netvlad}
Relja Arandjelovic, Petr Gronat, Akihiko Torii, Tomas Pajdla, and Josef Sivic.
\newblock Netvlad: Cnn architecture for weakly supervised place recognition.
\newblock In \emph{CVPR}, 2016.

\bibitem[Bai et~al.(2020)Bai, Luo, Zhou, Fu, Quan, and Tai]{bai2020d3feat}
Xuyang Bai, Zixin Luo, Lei Zhou, Hongbo Fu, Long Quan, and Chiew-Lan Tai.
\newblock D3feat: Joint learning of dense detection and description of 3d local features.
\newblock In \emph{CVPR}, 2020.

\bibitem[Besl and McKay(1992)]{besl1992method}
Paul~J Besl and Neil~D McKay.
\newblock Method for registration of 3-d shapes.
\newblock In \emph{Sensor fusion IV: control paradigms and data structures}, 1992.

\bibitem[Bosse et~al.(2003)Bosse, Newman, Leonard, Soika, Feiten, and Teller]{bosse2003atlas}
Michael Bosse, Paul Newman, John Leonard, Martin Soika, Wendelin Feiten, and Seth Teller.
\newblock An atlas framework for scalable mapping.
\newblock In \emph{ICRA}, 2003.

\bibitem[Br{\'e}gier(2021)]{bregier2021roma}
Romain Br{\'e}gier.
\newblock Deep regression on manifolds: a {3D} rotation case study.
\newblock 2021.

\bibitem[Campos et~al.(2021)Campos, Elvira, Rodr{\'\i}guez, Montiel, and Tard{\'o}s]{campos2021orb_slam3}
Carlos Campos, Richard Elvira, Juan J~G{\'o}mez Rodr{\'\i}guez, Jos{\'e}~MM Montiel, and Juan~D Tard{\'o}s.
\newblock Orb-slam3: An accurate open-source library for visual, visual--inertial, and multimap slam.
\newblock \emph{IEEE Transactions on Robotics}, 2021.

\bibitem[Cao et~al.(2018)Cao, Kobbelt, and Hu]{cao2018real}
Yan-Pei Cao, Leif Kobbelt, and Shi-Min Hu.
\newblock Real-time high-accuracy three-dimensional reconstruction with consumer rgb-d cameras.
\newblock \emph{ACM TOG}, 2018.

\bibitem[Chang et~al.(2024)Chang, Xu, Li, Chen, and Han]{chang2024gaussreg}
Jiahao Chang, Yinglin Xu, Yihao Li, Yuantao Chen, and Xiaoguang Han.
\newblock Gaussreg: Fast 3d registration with gaussian splatting.
\newblock In \emph{ECCV}, 2024.

\bibitem[Chen et~al.(2013)Chen, Bautembach, and Izadi]{chen2013scalable}
Jiawen Chen, Dennis Bautembach, and Shahram Izadi.
\newblock Scalable real-time volumetric surface reconstruction.
\newblock \emph{ACM TOG}, 2013.

\bibitem[Chen and Lee(2023)]{chen2023dreg}
Yu Chen and Gim~Hee Lee.
\newblock Dreg-nerf: Deep registration for neural radiance fields.
\newblock In \emph{ICCV}, 2023.

\bibitem[Cho et~al.(2021)Cho, Jo, and Kim]{cho2021sp}
Hae~Min Cho, HyungGi Jo, and Euntai Kim.
\newblock Sp-slam: Surfel-point simultaneous localization and mapping.
\newblock \emph{IEEE/ASME Transactions on Mechatronics}, 2021.

\bibitem[Choi et~al.(2015)Choi, Zhou, and Koltun]{choi2015robust}
Sungjoon Choi, Qian-Yi Zhou, and Vladlen Koltun.
\newblock Robust reconstruction of indoor scenes.
\newblock In \emph{CVPR}, 2015.

\bibitem[Choy et~al.(2019)Choy, Park, and Koltun]{Choy2019FCGF}
Christopher Choy, Jaesik Park, and Vladlen Koltun.
\newblock Fully convolutional geometric features.
\newblock In \emph{ICCV}, 2019.

\bibitem[Chung et~al.(2022)Chung, Tseng, Hsu, Shi, Hua, Yeh, Chen, Chen, and Hsu]{chung2022orbeez}
Chi-Ming Chung, Yang-Che Tseng, Ya-Ching Hsu, Xiang-Qian Shi, Yun-Hung Hua, Jia-Fong Yeh, Wen-Chin Chen, Yi-Ting Chen, and Winston~H Hsu.
\newblock Orbeez-slam: A real-time monocular visual slam with orb features and nerf-realized mapping.
\newblock \emph{arXiv preprint arXiv:2209.13274}, 2022.

\bibitem[Curless and Levoy(1996)]{curless1996volumetric}
Brian Curless and Marc Levoy.
\newblock {Volumetric method for building complex models from range images}.
\newblock In \emph{SIGGRAPH Conference on Computer Graphics}, 1996.

\bibitem[Dai et~al.(2017{\natexlab{a}})Dai, Chang, Savva, Halber, Funkhouser, and Nie{\ss}ner]{dai2017scannet}
Angela Dai, Angel~X Chang, Manolis Savva, Maciej Halber, Thomas Funkhouser, and Matthias Nie{\ss}ner.
\newblock Scannet: Richly-annotated 3d reconstructions of indoor scenes.
\newblock In \emph{CVPR}, 2017{\natexlab{a}}.

\bibitem[Dai et~al.(2017{\natexlab{b}})Dai, Nie{\ss}ner, Zollh{\"o}fer, Izadi, and Theobalt]{dai2017bundlefusion}
Angela Dai, Matthias Nie{\ss}ner, Michael Zollh{\"o}fer, Shahram Izadi, and Christian Theobalt.
\newblock Bundlefusion: Real-time globally consistent 3d reconstruction using on-the-fly surface reintegration.
\newblock \emph{ACM TOG}, 2017{\natexlab{b}}.

\bibitem[Endres et~al.(2012)Endres, Hess, Engelhard, Sturm, Cremers, and Burgard]{endres2012evaluation}
Felix Endres, J{\"u}rgen Hess, Nikolas Engelhard, J{\"u}rgen Sturm, Daniel Cremers, and Wolfram Burgard.
\newblock An evaluation of the rgb-d slam system.
\newblock In \emph{ICRA}, 2012.

\bibitem[Engel et~al.(2014)Engel, Sch{\"o}ps, and Cremers]{engel2014lsd}
Jakob Engel, Thomas Sch{\"o}ps, and Daniel Cremers.
\newblock Lsd-slam: Large-scale direct monocular slam.
\newblock In \emph{ECCV}, 2014.

\bibitem[Engelmann et~al.(2024)Engelmann, Manhardt, Niemeyer, Tateno, Pollefeys, and Tombari]{engelmann2024opennerf}
Francis Engelmann, Fabian Manhardt, Michael Niemeyer, Keisuke Tateno, Marc Pollefeys, and Federico Tombari.
\newblock Opennerf: Open set 3d neural scene segmentation with pixel-wise features and rendered novel views.
\newblock \emph{arXiv preprint arXiv:2404.03650}, 2024.

\bibitem[Fioraio et~al.(2015)Fioraio, Taylor, Fitzgibbon, Di~Stefano, and Izadi]{fioraio2015large}
Nicola Fioraio, Jonathan Taylor, Andrew Fitzgibbon, Luigi Di~Stefano, and Shahram Izadi.
\newblock Large-scale and drift-free surface reconstruction using online subvolume registration.
\newblock In \emph{CVPR}, 2015.

\bibitem[Gojcic et~al.(2019)Gojcic, Zhou, Wegner, and Wieser]{gojcic2019perfect}
Zan Gojcic, Caifa Zhou, Jan~D Wegner, and Andreas Wieser.
\newblock The perfect match: 3d point cloud matching with smoothed densities.
\newblock In \emph{CVPR}, 2019.

\bibitem[Goli et~al.(2023)Goli, Rebain, Sabour, Garg, and Tagliasacchi]{goli2023nerf2nerf}
Lily Goli, Daniel Rebain, Sara Sabour, Animesh Garg, and Andrea Tagliasacchi.
\newblock nerf2nerf: Pairwise registration of neural radiance fields.
\newblock In \emph{ICRA}, 2023.

\bibitem[Gu{\'e}don and Lepetit(2024)]{guedon2023sugar}
Antoine Gu{\'e}don and Vincent Lepetit.
\newblock Sugar: Surface-aligned gaussian splatting for efficient 3d mesh reconstruction and high-quality mesh rendering.
\newblock \emph{CVPR}, 2024.

\bibitem[Henry et~al.(2012)Henry, Krainin, Herbst, Ren, and Fox]{henry2012rgb}
Peter Henry, Michael Krainin, Evan Herbst, Xiaofeng Ren, and Dieter Fox.
\newblock Rgb-d mapping: Using kinect-style depth cameras for dense 3d modeling of indoor environments.
\newblock \emph{The International Journal of Robotics Research}, 2012.

\bibitem[Henry et~al.(2013)Henry, Fox, Bhowmik, and Mongia]{henry2013patch}
Peter Henry, Dieter Fox, Achintya Bhowmik, and Rajiv Mongia.
\newblock Patch volumes: Segmentation-based consistent mapping with rgb-d cameras.
\newblock In \emph{3DV}, 2013.

\bibitem[Hu et~al.(2023)Hu, Mao, Bao, Zhang, and Cui]{hu2023cp}
Jiarui Hu, Mao Mao, Hujun Bao, Guofeng Zhang, and Zhaopeng Cui.
\newblock {CP}-{SLAM}: Collaborative neural point-based {SLAM} system.
\newblock In \emph{NeurIPS}, 2023.

\bibitem[Huang et~al.(2024{\natexlab{a}})Huang, Yu, Chen, Geiger, and Gao]{Huang2DGS2024}
Binbin Huang, Zehao Yu, Anpei Chen, Andreas Geiger, and Shenghua Gao.
\newblock 2d gaussian splatting for geometrically accurate radiance fields.
\newblock In \emph{SIGGRAPH}, 2024{\natexlab{a}}.

\bibitem[Huang et~al.(2024{\natexlab{b}})Huang, Li, Cheng, and Yeung]{huang2024photo}
Huajian Huang, Longwei Li, Hui Cheng, and Sai-Kit Yeung.
\newblock Photo-slam: Real-time simultaneous localization and photorealistic mapping for monocular stereo and rgb-d cameras.
\newblock In \emph{CVPR}, 2024{\natexlab{b}}.

\bibitem[Huang et~al.(2021{\natexlab{a}})Huang, Huang, Song, and Hu]{huang_difusion}
Jiahui Huang, Shi-Sheng Huang, Haoxuan Song, and Shi-Min Hu.
\newblock Di-fusion: Online implicit 3d reconstruction with deep priors.
\newblock In \emph{CVPR}, 2021{\natexlab{a}}.

\bibitem[Huang et~al.(2021{\natexlab{b}})Huang, Gojcic, Usvyatsov, Wieser, and Schindler]{huang2021predator}
Shengyu Huang, Zan Gojcic, Mikhail Usvyatsov, Andreas Wieser, and Konrad Schindler.
\newblock Predator: Registration of 3d point clouds with low overlap.
\newblock In \emph{CVPR}, 2021{\natexlab{b}}.

\bibitem[Johari et~al.(2023)Johari, Carta, and Fleuret]{eslam_cvpr23}
M.~M. Johari, C. Carta, and F. Fleuret.
\newblock {ESLAM}: Efficient dense slam system based on hybrid representation of signed distance fields.
\newblock In \emph{CVPR}, 2023.

\bibitem[K{\"{a}}hler et~al.(2015)K{\"{a}}hler, Prisacariu, Ren, Sun, Torr, and Murray]{Kahler2015infiniTAM}
Olaf K{\"{a}}hler, Victor~Adrian Prisacariu, Carl~Yuheng Ren, Xin Sun, Philip H.~S. Torr, and David~William Murray.
\newblock Very high frame rate volumetric integration of depth images on mobile devices.
\newblock \emph{{IEEE} Trans. Vis. Comput. Graph.}, 2015.

\bibitem[K{\"a}hler et~al.(2016)K{\"a}hler, Prisacariu, and Murray]{kahler2016real}
Olaf K{\"a}hler, Victor~A Prisacariu, and David~W Murray.
\newblock Real-time large-scale dense 3d reconstruction with loop closure.
\newblock In \emph{ECCV}, 2016.

\bibitem[Keetha et~al.(2024)Keetha, Karhade, Jatavallabhula, Yang, Scherer, Ramanan, and Luiten]{keetha2023splatam}
Nikhil Keetha, Jay Karhade, Krishna~Murthy Jatavallabhula, Gengshan Yang, Sebastian Scherer, Deva Ramanan, and Jonathon Luiten.
\newblock Splatam: Splat, track \& map 3d gaussians for dense rgb-d slam.
\newblock \emph{CVPR}, 2024.

\bibitem[Keller et~al.(2013)Keller, Lefloch, Lambers, Izadi, Weyrich, and Kolb]{keller2013real}
Maik Keller, Damien Lefloch, Martin Lambers, Shahram Izadi, Tim Weyrich, and Andreas Kolb.
\newblock Real-time 3d reconstruction in dynamic scenes using point-based fusion.
\newblock In \emph{International Conference on 3D Vision (3DV)}, 2013.

\bibitem[Kerbl et~al.(2023)Kerbl, Kopanas, Leimk{\"u}hler, and Drettakis]{kerbl20233d}
Bernhard Kerbl, Georgios Kopanas, Thomas Leimk{\"u}hler, and George Drettakis.
\newblock 3d gaussian splatting for real-time radiance field rendering.
\newblock \emph{ACM TOG}, 2023.

\bibitem[Kerl et~al.(2013)Kerl, Sturm, and Cremers]{kerl2013dense}
Christian Kerl, J{\"u}rgen Sturm, and Daniel Cremers.
\newblock Dense visual slam for rgb-d cameras.
\newblock In \emph{IROS}, 2013.

\bibitem[Liso et~al.(2024)Liso, Sandström, Yugay, Gool, and Oswald]{liso2024loopyslam}
Lorenzo Liso, Erik Sandström, Vladimir Yugay, Luc~Van Gool, and Martin~R. Oswald.
\newblock Loopy-slam: Dense neural slam with loop closures.
\newblock In \emph{CVPR}, 2024.

\bibitem[Liu et~al.(2020)Liu, Gu, Zaw~Lin, Chua, and Theobalt]{liu2020neural}
Lingjie Liu, Jiatao Gu, Kyaw Zaw~Lin, Tat-Seng Chua, and Christian Theobalt.
\newblock Neural sparse voxel fields.
\newblock \emph{NeurIPS}, 2020.

\bibitem[Lorensen and Cline(1987)]{lorensen1987marching}
William~E Lorensen and Harvey~E Cline.
\newblock Marching cubes: A high resolution 3d surface construction algorithm.
\newblock \emph{ACM siggraph computer graphics}, 1987.

\bibitem[Mahdi~Johari et~al.(2022)Mahdi~Johari, Carta, and Fleuret]{mahdi2022eslam}
Mohammad Mahdi~Johari, Camilla Carta, and Fran{\c{c}}ois Fleuret.
\newblock Eslam: Efficient dense slam system based on hybrid representation of signed distance fields.
\newblock pages arXiv--2211, 2022.

\bibitem[Maier et~al.(2014)Maier, Sturm, and Cremers]{maier2014submap}
Robert Maier, J{\"u}rgen Sturm, and Daniel Cremers.
\newblock Submap-based bundle adjustment for 3d reconstruction from rgb-d data.
\newblock In \emph{Pattern Recognition: 36th German Conference, GCPR 2014, M{\"u}nster, Germany, September 2-5, 2014, Proceedings 36}, 2014.

\bibitem[Maier et~al.(2017)Maier, Schaller, and Cremers]{maier2017efficient}
R Maier, R Schaller, and D Cremers.
\newblock Efficient online surface correction for real-time large-scale 3d reconstruction. arxiv 2017.
\newblock \emph{arXiv preprint arXiv:1709.03763}, 2017.

\bibitem[Mao et~al.(2023)Mao, Yu, Wang, Wang, Xiong, and Liao]{mao2023ngel}
Yunxuan Mao, Xuan Yu, Kai Wang, Yue Wang, Rong Xiong, and Yiyi Liao.
\newblock Ngel-slam: Neural implicit representation-based global consistent low-latency slam system.
\newblock \emph{arXiv preprint arXiv:2311.09525}, 2023.

\bibitem[Marniok et~al.(2017)Marniok, Johannsen, and Goldluecke]{marniok2017efficient}
Nico Marniok, Ole Johannsen, and Bastian Goldluecke.
\newblock An efficient octree design for local variational range image fusion.
\newblock In \emph{German Conference on Pattern Recognition (GCPR)}, 2017.

\bibitem[Matsuki et~al.(2023{\natexlab{a}})Matsuki, Murai, Kelly, and Davison]{matsuki2023monogs}
Hidenobu Matsuki, Riku Murai, Paul~HJ Kelly, and Andrew~J Davison.
\newblock Gaussian splatting slam.
\newblock \emph{arXiv preprint arXiv:2312.06741}, 2023{\natexlab{a}}.

\bibitem[Matsuki et~al.(2023{\natexlab{b}})Matsuki, Tateno, Niemeyer, and Tombari]{matsuki2023newton}
Hidenobu Matsuki, Keisuke Tateno, Michael Niemeyer, and Federic Tombari.
\newblock Newton: Neural view-centric mapping for on-the-fly large-scale slam.
\newblock \emph{arXiv preprint arXiv:2303.13654}, 2023{\natexlab{b}}.

\bibitem[Matsuki et~al.(2024)Matsuki, Murai, Kelly, and Davison]{matsuki2023gaussian}
Hidenobu Matsuki, Riku Murai, Paul~HJ Kelly, and Andrew~J Davison.
\newblock Gaussian splatting slam.
\newblock \emph{CVPR}, 2024.

\bibitem[Mildenhall et~al.(2021)Mildenhall, Srinivasan, Tancik, Barron, Ramamoorthi, and Ng]{mildenhall2021nerf}
Ben Mildenhall, Pratul~P Srinivasan, Matthew Tancik, Jonathan~T Barron, Ravi Ramamoorthi, and Ren Ng.
\newblock Nerf: Representing scenes as neural radiance fields for view synthesis.
\newblock In \emph{ECCV}, 2021.

\bibitem[Mur-Artal and Tardos(2017)]{Mur-Artal2017ORB-SLAM2:Cameras}
Raul Mur-Artal and Juan~D. Tardos.
\newblock {ORB-SLAM2: An Open-Source SLAM System for Monocular, Stereo, and RGB-D Cameras}.
\newblock \emph{IEEE Transactions on Robotics}, 2017.

\bibitem[Newcombe et~al.(2011)Newcombe, Izadi, Hilliges, Molyneaux, Kim, Davison, Kohli, Shotton, Hodges, and Fitzgibbon]{newcombe2011kinectfusion}
Richard~A Newcombe, Shahram Izadi, Otmar Hilliges, David Molyneaux, David Kim, Andrew~J Davison, Pushmeet Kohli, Jamie Shotton, Steve Hodges, and Andrew~W Fitzgibbon.
\newblock Kinectfusion: Real-time dense surface mapping and tracking.
\newblock In \emph{ISMAR}, 2011.

\bibitem[Nießner et~al.(2013)Nießner, Zollhöfer, Izadi, and Stamminger]{niessner2013voxel_hashing}
Matthias Nießner, Michael Zollhöfer, Shahram Izadi, and Marc Stamminger.
\newblock Real-time 3d reconstruction at scale using voxel hashing.
\newblock \emph{ACM TOG}, 2013.

\bibitem[Oleynikova et~al.(2017)Oleynikova, Taylor, Fehr, Siegwart, and Nieto]{Oleynikova2017voxblox}
Helen Oleynikova, Zachary Taylor, Marius Fehr, Roland Siegwart, and Juan~I. Nieto.
\newblock Voxblox: Incremental 3d euclidean signed distance fields for on-board {MAV} planning.
\newblock In \emph{IROS}, 2017.

\bibitem[Pan et~al.(2024)Pan, Zhong, Wiesmann, Posewsky, Behley, and Stachniss]{pan2024pin_slam}
Y. Pan, X. Zhong, L. Wiesmann, T. Posewsky, J. Behley, and C. Stachniss.
\newblock {PIN-SLAM: LiDAR SLAM Using a Point-Based Implicit Neural Representation for Achieving Global Map Consistency}.
\newblock \emph{IEEE Transactions on Robotics (TRO)}, 2024.

\bibitem[Park et~al.(2017)Park, Zhou, and Koltun]{park2017colored}
Jaesik Park, Qian-Yi Zhou, and Vladlen Koltun.
\newblock Colored point cloud registration revisited.
\newblock In \emph{ICCV}, pages 143--152, 2017.

\bibitem[Peretroukhin et~al.(2020)Peretroukhin, Giamou, Rosen, Greene, Roy, and Kelly]{peretroukhin2020rotation_averaging}
Valentin Peretroukhin, Matthew Giamou, David~M Rosen, W~Nicholas Greene, Nicholas Roy, and Jonathan Kelly.
\newblock A smooth representation of belief over so (3) for deep rotation learning with uncertainty.
\newblock \emph{arXiv preprint arXiv:2006.01031}, 2020.

\bibitem[Qin et~al.(2023)Qin, Yu, Wang, Guo, Peng, Ilic, Hu, and Xu]{qin2023geotransformer}
Zheng Qin, Hao Yu, Changjian Wang, Yulan Guo, Yuxing Peng, Slobodan Ilic, Dewen Hu, and Kai Xu.
\newblock Geotransformer: Fast and robust point cloud registration with geometric transformer.
\newblock \emph{IEEE TPAMI}, 2023.

\bibitem[Reijgwart et~al.(2019)Reijgwart, Millane, Oleynikova, Siegwart, Cadena, and Nieto]{reijgwart2019voxgraph}
Victor Reijgwart, Alexander Millane, Helen Oleynikova, Roland Siegwart, Cesar Cadena, and Juan Nieto.
\newblock Voxgraph: Globally consistent, volumetric mapping using signed distance function submaps.
\newblock \emph{IEEE Robotics and Automation Letters}, 2019.

\bibitem[Rosinol et~al.(2022)Rosinol, Leonard, and Carlone]{Rosinol2022NeRF-SLAM:Fields}
Antoni Rosinol, John~J. Leonard, and Luca Carlone.
\newblock {NeRF-SLAM: Real-Time Dense Monocular SLAM with Neural Radiance Fields}.
\newblock \emph{arXiv}, 2022.

\bibitem[Rusu et~al.(2009)Rusu, Blodow, and Beetz]{rusu2009fpfh}
Radu~Bogdan Rusu, Nico Blodow, and Michael Beetz.
\newblock Fast point feature histograms ({FPFH}) for 3d registration.
\newblock In \emph{ICRA}, 2009.

\bibitem[Sandstr{\"o}m et~al.(2023{\natexlab{a}})Sandstr{\"o}m, Li, Van~Gool, and Oswald]{sandstrom2023point}
Erik Sandstr{\"o}m, Yue Li, Luc Van~Gool, and Martin~R Oswald.
\newblock Point-slam: Dense neural point cloud-based slam.
\newblock In \emph{ICCV}, 2023{\natexlab{a}}.

\bibitem[Sandstr{\"o}m et~al.(2023{\natexlab{b}})Sandstr{\"o}m, Ta, Van~Gool, and Oswald]{sandstrom2023uncle}
Erik Sandstr{\"o}m, Kevin Ta, Luc Van~Gool, and Martin~R Oswald.
\newblock Uncle-slam: Uncertainty learning for dense neural slam.
\newblock In \emph{Int. Conf. Comput. Vis. Worksh.}, 2023{\natexlab{b}}.

\bibitem[Sandstr{\"o}m et~al.(2024)Sandstr{\"o}m, Tateno, Oechsle, Niemeyer, Van~Gool, Oswald, and Tombari]{sandstrom2024splat}
Erik Sandstr{\"o}m, Keisuke Tateno, Michael Oechsle, Michael Niemeyer, Luc Van~Gool, Martin~R Oswald, and Federico Tombari.
\newblock Splat-slam: Globally optimized rgb-only slam with 3d gaussians.
\newblock \emph{arXiv preprint arXiv:2405.16544}, 2024.

\bibitem[Sarlin et~al.(2019)Sarlin, Cadena, Siegwart, and Dymczyk]{sarlin2019coarse}
Paul-Edouard Sarlin, Cesar Cadena, Roland Siegwart, and Marcin Dymczyk.
\newblock From coarse to fine: Robust hierarchical localization at large scale.
\newblock In \emph{CVPR}, 2019.

\bibitem[Sch\"{o}nberger and Frahm(2016)]{schoenberger2016colmap}
Johannes~Lutz Sch\"{o}nberger and Jan-Michael Frahm.
\newblock Structure-from-motion revisited.
\newblock In \emph{CVPR}, 2016.

\bibitem[Schops et~al.(2019)Schops, Sattler, and Pollefeys]{schops2019bad}
Thomas Schops, Torsten Sattler, and Marc Pollefeys.
\newblock {BAD SLAM}: Bundle adjusted direct {RGB-D} {SLAM}.
\newblock In \emph{CVPR}, 2019.

\bibitem[Steinbrucker et~al.(2013)Steinbrucker, Kerl, and Cremers]{steinbrucker2013large}
Frank Steinbrucker, Christian Kerl, and Daniel Cremers.
\newblock Large-scale multi-resolution surface reconstruction from rgb-d sequences.
\newblock In \emph{ICCV}, 2013.

\bibitem[Straub et~al.(2019{\natexlab{a}})Straub, Whelan, Ma, Chen, Wijmans, Green, Engel, Mur-Artal, Ren, Verma, Clarkson, Yan, Budge, Yan, Pan, Yon, Zou, Leon, Carter, Briales, Gillingham, Mueggler, Pesqueira, Savva, Batra, Strasdat, Nardi, Goesele, Lovegrove, and Newcombe]{replica19arxiv}
Julian Straub, Thomas Whelan, Lingni Ma, Yufan Chen, Erik Wijmans, Simon Green, Jakob~J. Engel, Raul Mur-Artal, Carl Ren, Shobhit Verma, Anton Clarkson, Mingfei Yan, Brian Budge, Yajie Yan, Xiaqing Pan, June Yon, Yuyang Zou, Kimberly Leon, Nigel Carter, Jesus Briales, Tyler Gillingham, Elias Mueggler, Luis Pesqueira, Manolis Savva, Dhruv Batra, Hauke~M. Strasdat, Renzo~De Nardi, Michael Goesele, Steven Lovegrove, and Richard Newcombe.
\newblock The {R}eplica dataset: A digital replica of indoor spaces.
\newblock \emph{arXiv preprint arXiv:1906.05797}, 2019{\natexlab{a}}.

\bibitem[Straub et~al.(2019{\natexlab{b}})Straub, Whelan, Ma, Chen, Wijmans, Green, Engel, Mur-Artal, Ren, Verma, et~al.]{straub2019replica}
Julian Straub, Thomas Whelan, Lingni Ma, Yufan Chen, Erik Wijmans, Simon Green, Jakob~J Engel, Raul Mur-Artal, Carl Ren, Shobhit Verma, et~al.
\newblock The replica dataset: A digital replica of indoor spaces.
\newblock \emph{arXiv preprint arXiv:1906.05797}, 2019{\natexlab{b}}.

\bibitem[St{\"u}ckler and Behnke(2014)]{stuckler2014multi}
J{\"o}rg St{\"u}ckler and Sven Behnke.
\newblock Multi-resolution surfel maps for efficient dense 3d modeling and tracking.
\newblock \emph{Journal of Visual Communication and Image Representation}, 2014.

\bibitem[Sturm et~al.(2012{\natexlab{a}})Sturm, Engelhard, Endres, Burgard, and Cremers]{Sturm2012ASystems}
Jürgen Sturm, Nikolas Engelhard, Felix Endres, Wolfram Burgard, and Daniel Cremers.
\newblock {A benchmark for the evaluation of RGB-D SLAM systems}.
\newblock In \emph{International Conference on Intelligent Robots and Systems (IROS)}, 2012{\natexlab{a}}.

\bibitem[Sturm et~al.(2012{\natexlab{b}})Sturm, Engelhard, Endres, Burgard, and Cremers]{sturm2012benchmark}
J{\"u}rgen Sturm, Nikolas Engelhard, Felix Endres, Wolfram Burgard, and Daniel Cremers.
\newblock A benchmark for the evaluation of rgb-d slam systems.
\newblock In \emph{IROS}, 2012{\natexlab{b}}.

\bibitem[Sucar et~al.(2021{\natexlab{a}})Sucar, Liu, Ortiz, and Davison]{Sucar2021IMAP:Real-Time}
Edgar Sucar, Shikun Liu, Joseph Ortiz, and Andrew~J. Davison.
\newblock {iMAP: Implicit Mapping and Positioning in Real-Time}.
\newblock In \emph{ICCV}, 2021{\natexlab{a}}.

\bibitem[Sucar et~al.(2021{\natexlab{b}})Sucar, Liu, Ortiz, and Davison]{sucar2021imap}
Edgar Sucar, Shikun Liu, Joseph Ortiz, and Andrew~J Davison.
\newblock imap: Implicit mapping and positioning in real-time.
\newblock In \emph{ICCV}, 2021{\natexlab{b}}.

\bibitem[Tang et~al.(2023{\natexlab{a}})Tang, Zhang, Yu, Wang, and Xu]{mipsfusion_siga23}
Yijie Tang, Jiazhao Zhang, Zhinan Yu, He Wang, and Kai Xu.
\newblock Mips-fusion: Multi-implicit-submaps for scalable and robust online neural rgb-d reconstruction.
\newblock \emph{ACM TOG}, 2023{\natexlab{a}}.

\bibitem[Tang et~al.(2023{\natexlab{b}})Tang, Zhang, Yu, Wang, and Xu]{tang2023mips}
Yijie Tang, Jiazhao Zhang, Zhinan Yu, He Wang, and Kai Xu.
\newblock Mips-fusion: Multi-implicit-submaps for scalable and robust online neural rgb-d reconstruction.
\newblock \emph{arXiv preprint arXiv:2308.08741}, 2023{\natexlab{b}}.

\bibitem[Teed and Deng(2021)]{teed2021droid}
Zachary Teed and Jia Deng.
\newblock Droid-slam: Deep visual slam for monocular, stereo, and rgb-d cameras.
\newblock \emph{NeurIPS}, 2021.

\bibitem[Tombari et~al.(2010)Tombari, Salti, and Di~Stefano]{tombari2010unique}
Federico Tombari, Samuele Salti, and Luigi Di~Stefano.
\newblock Unique signatures of histograms for local surface description.
\newblock In \emph{ECCV}, 2010.

\bibitem[Vaswani et~al.(2017)Vaswani, Shazeer, Parmar, Uszkoreit, Jones, Gomez, Kaiser, and Polosukhin]{vaswani2017attention}
Ashish Vaswani, Noam Shazeer, Niki Parmar, Jakob Uszkoreit, Llion Jones, Aidan~N Gomez, {\L}ukasz Kaiser, and Illia Polosukhin.
\newblock Attention is all you need.
\newblock In \emph{NeurIPS}, 2017.

\bibitem[Wang et~al.(2016)Wang, Wang, and Liang]{wang2016online}
Hao Wang, Jun Wang, and Wang Liang.
\newblock Online reconstruction of indoor scenes from rgb-d streams.
\newblock In \emph{CVPR}, 2016.

\bibitem[Wang et~al.(2023{\natexlab{a}})Wang, Wang, and Agapito]{wang2023co}
Hengyi Wang, Jingwen Wang, and Lourdes Agapito.
\newblock Co-slam: Joint coordinate and sparse parametric encodings for neural real-time slam.
\newblock In \emph{CVPR}, 2023{\natexlab{a}}.

\bibitem[Wang et~al.(2004)Wang, Bovik, Sheikh, and Simoncelli]{wang2004image}
Zhou Wang, Alan~C Bovik, Hamid~R Sheikh, and Eero~P Simoncelli.
\newblock Image quality assessment: from error visibility to structural similarity.
\newblock \emph{IEEE transactions on image processing}, 2004.

\bibitem[Wang et~al.(2023{\natexlab{b}})Wang, Shen, Gao, Huang, Munkberg, Hasselgren, Gojcic, Chen, and Fidler]{wang2023fegr}
Zian Wang, Tianchang Shen, Jun Gao, Shengyu Huang, Jacob Munkberg, Jon Hasselgren, Zan Gojcic, Wenzheng Chen, and Sanja Fidler.
\newblock Neural fields meet explicit geometric representations for inverse rendering of urban scenes.
\newblock In \emph{CVPR}, 2023{\natexlab{b}}.

\bibitem[Whelan et~al.(2012)Whelan, McDonald, Kaess, Fallon, Johannsson, and Leonard]{whelan2012kintinuous}
Thomas Whelan, John McDonald, Michael Kaess, Maurice Fallon, Hordur Johannsson, and John~J. Leonard.
\newblock Kintinuous: Spatially extended kinectfusion.
\newblock In \emph{Proceedings of RSS '12 Workshop on RGB-D: Advanced Reasoning with Depth Cameras}, 2012.

\bibitem[Whelan et~al.(2015)Whelan, Leutenegger, Salas-Moreno, Glocker, and Davison]{whelan2015elasticfusion}
Thomas Whelan, Stefan Leutenegger, Renato Salas-Moreno, Ben Glocker, and Andrew Davison.
\newblock Elasticfusion: Dense slam without a pose graph.
\newblock In \emph{Robotics: Science and Systems (RSS)}, 2015.

\bibitem[Yan et~al.(2024)Yan, Qu, Xu, Zhao, Wang, Wang, and Li]{yan2024gs}
Chi Yan, Delin Qu, Dan Xu, Bin Zhao, Zhigang Wang, Dong Wang, and Xuelong Li.
\newblock Gs-slam: Dense visual slam with 3d gaussian splatting.
\newblock In \emph{CVPR}, 2024.

\bibitem[Yan et~al.(2017)Yan, Ye, and Ren]{yan2017dense}
Zhixin Yan, Mao Ye, and Liu Ren.
\newblock Dense visual slam with probabilistic surfel map.
\newblock \emph{IEEE TVCG}, 2017.

\bibitem[Yang et~al.(2022{\natexlab{a}})Yang, Li, Zhai, Ming, Liu, and Zhang]{yang2022vox}
Xingrui Yang, Hai Li, Hongjia Zhai, Yuhang Ming, Yuqian Liu, and Guofeng Zhang.
\newblock Vox-fusion: Dense tracking and mapping with voxel-based neural implicit representation.
\newblock In \emph{IEEE International Symposium on Mixed and Augmented Reality (ISMAR)}, 2022{\natexlab{a}}.

\bibitem[Yang et~al.(2022{\natexlab{b}})Yang, Ming, Cui, and Calway]{yang2022fd}
Xingrui Yang, Yuhang Ming, Zhaopeng Cui, and Andrew Calway.
\newblock Fd-slam: 3-d reconstruction using features and dense matching.
\newblock In \emph{2022 International Conference on Robotics and Automation (ICRA)}, 2022{\natexlab{b}}.

\bibitem[Yen-Chen et~al.(2021)Yen-Chen, Florence, Barron, Rodriguez, Isola, and Lin]{yen2020inerf}
Lin Yen-Chen, Pete Florence, Jonathan~T. Barron, Alberto Rodriguez, Phillip Isola, and Tsung-Yi Lin.
\newblock {iNeRF}: Inverting neural radiance fields for pose estimation.
\newblock In \emph{({IROS})}, 2021.

\bibitem[Yeshwanth et~al.(2023)Yeshwanth, Liu, Nie{\ss}ner, and Dai]{yeshwanthliu2023scannetpp}
Chandan Yeshwanth, Yueh-Cheng Liu, Matthias Nie{\ss}ner, and Angela Dai.
\newblock Scannet++: A high-fidelity dataset of 3d indoor scenes.
\newblock In \emph{ICCV}, 2023.

\bibitem[Yu et~al.(2024)Yu, Sattler, and Geiger]{Yu2024GOF}
Zehao Yu, Torsten Sattler, and Andreas Geiger.
\newblock Gaussian opacity fields: Efficient high-quality compact surface reconstruction in unbounded scenes.
\newblock \emph{arXiv:2404.10772}, 2024.

\bibitem[Yugay et~al.(2023)Yugay, Li, Gevers, and Oswald]{yugay2023gaussian}
Vladimir Yugay, Yue Li, Theo Gevers, and Martin~R Oswald.
\newblock Gaussian-slam: Photo-realistic dense slam with gaussian splatting.
\newblock \emph{arXiv preprint arXiv:2312.10070}, 2023.

\bibitem[Zeng et~al.(2017)Zeng, Song, Nie{\ss}ner, Fisher, Xiao, and Funkhouser]{zeng20173dmatch}
Andy Zeng, Shuran Song, Matthias Nie{\ss}ner, Matthew Fisher, Jianxiong Xiao, and Thomas Funkhouser.
\newblock 3dmatch: Learning local geometric descriptors from rgb-d reconstructions.
\newblock In \emph{CVPR}, 2017.

\bibitem[Zhang et~al.(2024{\natexlab{a}})Zhang, Fang, Shrestha, Liang, Long, and Tan]{zhang2024rade}
Baowen Zhang, Chuan Fang, Rakesh Shrestha, Yixun Liang, Xiaoxiao Long, and Ping Tan.
\newblock Rade-gs: Rasterizing depth in gaussian splatting.
\newblock \emph{arXiv preprint arXiv:2406.01467}, 2024{\natexlab{a}}.

\bibitem[Zhang et~al.(2024{\natexlab{b}})Zhang, Sandstr{\"o}m, Zhang, Patel, Van~Gool, and Oswald]{zhang2024glorie}
Ganlin Zhang, Erik Sandstr{\"o}m, Youmin Zhang, Manthan Patel, Luc Van~Gool, and Martin~R Oswald.
\newblock Glorie-slam: Globally optimized rgb-only implicit encoding point cloud slam.
\newblock \emph{arXiv preprint arXiv:2403.19549}, 2024{\natexlab{b}}.

\bibitem[Zhang et~al.(2020)Zhang, Chen, Wang, Wang, and Sun]{zhang2020dense}
Heng Zhang, Guodong Chen, Zheng Wang, Zhenhua Wang, and Lining Sun.
\newblock Dense 3d mapping for indoor environment based on feature-point slam method.
\newblock In \emph{2020 the 4th International Conference on Innovation in Artificial Intelligence}, 2020.

\bibitem[Zhang et~al.(2018)Zhang, Isola, Efros, Shechtman, and Wang]{zhang2018unreasonable}
Richard Zhang, Phillip Isola, Alexei~A Efros, Eli Shechtman, and Oliver Wang.
\newblock The unreasonable effectiveness of deep features as a perceptual metric.
\newblock In \emph{CVPR}, 2018.

\bibitem[Zhang et~al.(2023)Zhang, Tosi, Mattoccia, and Poggi]{zhang2023goslam}
Youmin Zhang, Fabio Tosi, Stefano Mattoccia, and Matteo Poggi.
\newblock Go-slam: Global optimization for consistent 3d instant reconstruction.
\newblock In \emph{ICCV}, 2023.

\bibitem[Zhou et~al.(2016)Zhou, Park, and Koltun]{zhou2016fgr}
Qian-Yi Zhou, Jaesik Park, and Vladlen Koltun.
\newblock Fast global registration.
\newblock In \emph{ECCV}, 2016.

\bibitem[Zhu et~al.(2022)Zhu, Peng, Larsson, Xu, Bao, Cui, Oswald, and Pollefeys]{zhu2022nice}
Zihan Zhu, Songyou Peng, Viktor Larsson, Weiwei Xu, Hujun Bao, Zhaopeng Cui, Martin~R Oswald, and Marc Pollefeys.
\newblock Nice-slam: Neural implicit scalable encoding for slam.
\newblock In \emph{CVPR}, 2022.

\end{thebibliography}
}

\clearpage

\setcounter{page}{1}
\setcounter{section}{0}
\counterwithin{figure}{section} 
\counterwithin{table}{section}
\renewcommand{\thesection}{\Alph{section}} 
\maketitlesupplementary

\begin{abstract}
    This supplementary material includes a video of \ours running on a multi-room scene, showcasing the effectiveness of the online loop closure module of \ours. Furthermore, we provide the implementation details and statistics on loop closure and pose graph optimization (PGO). We also present more qualitative results and ablation studies. Lastly, we discuss the limitations and future work.
\end{abstract}

%
%
\section{Video}
We submit a video \texttt{loopsplat\_0054.mp4}, demonstrating \ours's online tracking and reconstruction capabilities on ScanNet~\cite{dai2017scannet}  \texttt{scene0054}. This video showcases the effectiveness of our globally consistent reconstruction process. The visualization includes the reconstructed mesh, the colorized camera trajectory that denotes average translation error from the ground truth trajectory -- see heatmap legend on the right, and the point cloud observed from the current frame colored in blue. As the camera completes its scan of the first room, one can clearly observe the significant improvements achieved through loop closure. While substantial drift occurs in the bathroom and storage room (the leftmost room), the online loop closure (LC) module in \ours successfully corrects the accumulated error when the camera revisits the first room at the end of the video. This correction highlights the robustness of our method in maintaining global consistency throughout the reconstruction process.

\section{Implementation Details} 
\label{sec:supp_details}

\boldparagraph{Hyperparameters.} \cref{tab:supp_configs} lists the hyperparameters used in our system, including $\lambda_c$ in the tracking loss, learning rates $l_r$ for rotation and $l_t$ for translation, and the number of optimization iterations $\text{iter}_t$ for tracking and $\text{iter}_m$ for mapping on the reported Replica~\cite{straub2019replica}, TUM-RGBD~\cite{Sturm2012ASystems}, ScanNet~\cite{dai2017scannet}, and ScanNet++~\cite{yeshwanthliu2023scannetpp} datasets. Additionally we set $\lambda_\text{color}$, $\lambda_\text{depth}$, and $\lambda_\text{reg}$ to 1 in the mapping loss $\mathcal{L}_\text{render}$ for all datasets.
\begin{table}[htb]
\centering
\scriptsize
\begin{tabularx}{\columnwidth}{lRRRR}
\toprule
\textbf{Params} & Replica & TUM-RGBD & ScanNet & ScanNet++ \\
\midrule
$\lambda_c$ & 0.95 & 0.6 & 0.6 & 0.5 \\
$l_r$ & 0.0002 & 0.002 & 0.002 & 0.002 \\
$l_t$ & 0.002 & 0.01 & 0.01 & 0.01  \\
$\text{iter}_t$ & 60 & 200 & 200 & 300\\
$\text{iter}_m$ & 100 & 100 & 100 & 500 \\
\bottomrule
\end{tabularx}
\caption{\textbf{Per-dataset Hyperparameters.}}
\label{tab:supp_configs}
\end{table}

\boldparagraph{Submap Initialization.} A new submap is triggered based on motion heuristics with the displacement threshold $d_\text{thre}=0.5~[m]$ and rotation threshold $\theta_\text{thre}=50^{\circ}$. For the ScanNet and ScanNet++ datasets, we adopted a different approach to submap initialization. Motion heuristics were not employed, primarily due to two factors: significant motion blur in ScanNet and substantial per-frame motion in ScanNet++ (\cf \cref{tab:datasets_avg_motion}). Instead, we implemented a fixed interval system for triggering new submaps. Specifically, we set intervals of 50 frames for ScanNet and 100 for ScanNet++.

\boldparagraph{Frame-to-model Tracking.}
Instead of estimating the current camera pose $\mathbf{T}_{j}$ directly, we optimize the relative camera pose $\mathbf{T}_{j-1, j}$ of frame $j$ with respect to $j-1$. To achieve the equivalent of rendering at the current pose $\mathbf{T}_j$, we transform the submap  with the relative transformation $\mathbf{T}_{j-1, j}^{-1}$ and render from the last camera pose $\mathbf{T}_{j-1}$ to get the rendered color $\hat{\mathbf{I}}_j$ and depth $\hat{\mathbf{D}}_j$.

\boldparagraph{Tracking Loss.} The inlier mask $M_\text{inlier}$ in the tracking loss filters out pixels with depth errors $50$ times larger than the median depth error of the current re-rendered depth map. Pixels without valid depth input are also excluded as the inconsistent re-rendering in those areas can hinder the pose optimization. For the soft alpha mask, we adopt $M_\text{alpha} = \alpha^3$ for per-pixel loss weighting. On the ScanNet++ dataset, if at the initialized pose the re-rendering loss is 50 times larger than the running average during tracking optimization, we use ICP odometry~\cite{park2017colored} to re-initialize the pose for the current frame.

\boldparagraph{Submap Expansion.} When selecting candidates to add to the submap at a new keyframe, we uniformly sample $M_k$ points from pixels that meet either the alpha value condition or the depth discrepancy condition. $M_k$ is set to $30K$ for TUM-RGBD and ScanNet datasets, $100K$ for ScanNet++, and all available points that meet either condition for Replica. The alpha threshold $\alpha_\text{thre}$ is set to 0.98 across all datasets. The depth discrepancy condition masks pixels where the depth error exceeds 40 times the median depth error of the current frame.

\boldparagraph{Submap Update.} The radius $\rho$ for the neighborhood check when adding new Gaussians to the submap is set to $1cm$. Newly added Gaussians are initialized with opacity values 0.5 and their initial scales are set to the nearest neighbor distances within the submap. As mentioned in the main paper, the Gaussians are not pruned until optimization finishes. After the mapping optimization for the new keyframe, we prune Gaussians that have opacity values lower than a threshold $o_\text{thre}$. We set $o_\text{thre}=0.1$ for Replica and $0.5$ for all other datasets. 

\boldparagraph{Submap Merging.}
Upon completing the mapping and tracking of all frames for the input sequence, we merge the saved submaps into a global map. The mesh is extracted by TSDF fusion~\cite{curless1996volumetric} using the rendered depth maps and estimated poses from the submaps. Then we use the reconstructed mesh vertices to initialize the Gaussian centers of the global map, providing a good starting point as they represent the scene geometry. We perform color refinement on the global map for $30K$ iterations using the same hyperparameters as in~\cite{kerbl20233d}. The Gaussian parameters of the global map are optimized from scratch using the RGB-D input and our estimated camera poses.

\begin{table}[htp]
\centering
\setlength{\tabcolsep}{15pt}
\renewcommand{\arraystretch}{1.2}
\resizebox{\columnwidth}{!}{
\begin{tabular}{lrrrr}
\toprule
\textbf{Params} & Replica & TUM RGB-D & ScanNet & ScanNet++ \\
\midrule
$\text{lr}_\text{rotation}$ & 0.003 & 0.015 & 0.015 & 0.015 \\
$\text{lr}_\text{translation}$ & 0.001 & 0.005 & 0.005 & 0.005  \\
$\text{lr}_\text{exposure}$ & 0.1 & 0.1 & 0.1 & 0.1\\
$\text{overlap}_\text{min}$ & 0.1 & 0.2 & 0.2 & 0.2 \\
$\text{interval}_\text{min}$ & 2 & 4 & 3 & 1\\
\bottomrule
\end{tabular}
}
\caption{\textbf{Per-dataset Hyperparameters on Loop Closure.}}
\label{tab:lc_configs}
\end{table}
\boldparagraph{Loop Detection.} 
For NetVLAD~\cite{arandjelovic2016netvlad}, we use the pretrained weights \texttt{VGG16-NetVLAD-Pitts30K} from HLoc~\cite{sarlin2019coarse}. 
We compute the cosine similarities of all keyframes within the $i$-th submap and determine the self-similarity score $s_\mathrm{self}^i$ corresponding to their $\mathrm{p}$-th percentile. We set $\mathrm{p}=50$ on Replica, TUM RGB-D, and ScanNet and $\mathrm{p}=33$ on ScanNet++. After getting the initial loops from the visual similarity between submaps, we further filter detected loops by computing their overlap ratio ($\mathrm{OR}$) using the pose estimated from the front-end. The overlapping region between $\mathbf{P}$ and $\mathbf{Q}$ is defined as in~\cite{huang2021predator}:
\begin{equation}
    \mathrm{OR}=\frac{1}{|\mathcal{K}_{ij}|} \sum_{(\mathbf{p},\mathbf{q})\in\mathcal{K}_{ij}} \left[ \left\|\mathbf{T}_\mathbf{P\rightarrow Q}(\mathbf{p}) - \mathbf{q}\right\|_2 < \tau_1 \right],
\end{equation}
with $\left[\cdot\right]$ the Iverson bracket and
$(\mathbf{p \in P},\mathbf{q \in Q})\in\mathcal{K}_{ij}$ the set of putative correspondences found by reciprocal matching the closest point between $\mathbf{P}$ and $\mathbf{Q}$. We set $\tau_1 = 0.1m$ on Replica and  $\tau_1 = 0.2m$ on TUM RGB-D, ScanNet, and ScanNet++. The selected thresholds are quite loose compared to standard point cloud registration, as we only need to ensure that two submaps have a spatial overlap for the next step. We also remove the loops where two submaps are too temporally close to each other to avoid redundant computations. We set the minimum submap id interval (interval$_{min}$) (\cf \cref{tab:lc_configs}) and remove the loop edges whose submap id distances are smaller than interval$_{min}$.

\boldparagraph{3DGS Registration.} We first find the overlapping viewpoints between two submaps using NetVLAD, as discussed in the main paper. For all datasets, we select the top-$k$ pairs as the overlapping viewpoints, $k=2$. In multi-view pose estimation, we optimize the camera pose parameters (\ie translation and rotation) and the exposure coefficients for selected viewpoints because the exposure of renders in different submaps can differ. We set different learning rates of parameters in \cref{tab:lc_configs}. The learning rates of camera pose parameters are significantly smaller because Replica is a synthetic dataset with high-quality RGB-D measurements from rendering; thus, the step size for optimization should be smaller. The learning rates on the three real-world datasets are consistent with each other. 

\boldparagraph{Number of LCs.} We report the number of frames, submaps, and loop closures (LCs) for each scene in our \ours system. On Replica scenes, LCs occur on average every 500 frames, about 4 times per scene (\cref{tab:replica_pgos}). The relatively low frequency of LCs in Replica is due to its single-room layouts and shorter sequences (approximately 2000 frames). In contrast, ScanNet~\cite{dai2017scannet} scenes feature longer sequences, averaging 4000 frames per scene (\cf \cref{tab:scanne_pgos}). More challenging scenes like \texttt{Scene 00, 54}, and \texttt{233} require \ours to create over 100 submaps and perform more than 30 pose graph optimizations (PGOs) per scene, which is attributed to their high sequence lengths. The TUM RGB-D dataset presents a mix of long and short sequences (\cf \cref{tab:tum_pgos}), resulting in varied numbers of submaps and PGOs across its scenes. This diversity in scene complexity and sequence length across datasets showcases the adaptability of \ours to different scene capturing scenarios.

\begin{table}[h]
\centering

\begin{subtable}[t]{\columnwidth}
\centering
\setlength{\tabcolsep}{4pt}
\renewcommand{\arraystretch}{1.2}
\resizebox{\columnwidth}{!}
{                
\begin{tabular}{lrrrrrrrrr}
\toprule
Method & \texttt{r0} & \texttt{r1} & \texttt{r2} & \texttt{o0}& \texttt{o1} & \texttt{o2} & \texttt{o3} &  \texttt{o4}  & Avg. \\ \midrule
$\#$ Frames & 2000   & 2000   & 2000  & 2000 & 2000 & 2000 & 2000   & 2000  & 2000\\
$\#$ Submaps & 38   & 25   & 33  & 27 & 11 & 39 & 45   & 39 & 32 \\
$\#$ LCs &  2 &  8  & 4 & 3 & 4  & 1   &2    & 6 & 4\\
\bottomrule 
\end{tabular}
}
\caption{Replica~\cite{replica19arxiv}}
\label{tab:replica_pgos}
\end{subtable}

\vspace{0.4cm}

\begin{subtable}[t]{\columnwidth}
\centering
\setlength{\tabcolsep}{4pt}
\renewcommand{\arraystretch}{1.05}
\resizebox{\columnwidth}{!}
{                
\begin{tabular}{lrrrrrrrrr}
\toprule
Method & \texttt{00}&  \texttt{54} & \texttt{59} & \texttt{106} & \texttt{169} & \texttt{181} & \texttt{207}  & \texttt{233}  & Avg. \\ \midrule
$\#$ Frames & 5578   & 6629 &1807   & 2324  & 2034 & 2349 & 1988   & 7643  & 4073  \\
$\#$ Submaps & 112   & 132   & 36  & 47 & 41 & 47 & 39   & 153  & 76\\
$\#$ LCs &  48  &  36  & 17 & 4 & 11  & 15   & 19  & 55   &  26\\
\bottomrule 
\end{tabular}
}
\caption{ScanNet~\cite{dai2017scannet}}
\label{tab:scanne_pgos}
\end{subtable}

\vspace{0.4cm}

\begin{subtable}[t]{\columnwidth}
\centering
\setlength{\tabcolsep}{4pt}
\renewcommand{\arraystretch}{1.2}
\resizebox{\columnwidth}{!}
{                
\begin{tabular}{lrrrrrr}
\toprule
Method & \texttt{fr1/desk1} & \texttt{fr1/desk2} & \texttt{fr1/room} & \texttt{fr2/xyz} & \texttt{fr3/office}   & Avg. \\ \midrule
$\#$ Frames & 595   & 640   & 1362  & 3669 & 2585 & 1770 \\
$\#$ Submaps & 14 &  15   & 24 & 6 & 39 & 20\\
$\#$ LCs &  7 &  7  & 6 & 2 & 5 & 5\\
\bottomrule 
\end{tabular}
}
\caption{TUM-RGBD~\cite{sturm2012benchmark}}
\label{tab:tum_pgos}
\end{subtable}

\vspace{0.2cm}
\caption{\textbf{Number of Submaps and PGOs Across Different Datasets}.}
\end{table}

\section{Datasets}
We first specify the ScanNet++ sequences used throughout our evaluation: (a) \texttt{b20a261fdf}, (b) \texttt{8b5caf3398}, (c) \texttt{fb05e13ad1}, (d) \texttt{2e74812d00}, (e) \texttt{281bc17764}. Some sudden large motions occur in the DSLR-captured sequences. To avoid this, we only use the first 250 frames of each sequence.
\cref{tab:datasets_avg_motion} shows the average ground truth frame translation distance and rotation degree per dataset on the scenes (and frame length) we evaluated. The average motion on ScanNet++ is about 10$\times$ larger than in other datasets, making it a challenging dataset for accurate pose estimation and, hence, highlighting the robustness of \ours given its superior performance on it.

\begin{table}[h]
\centering
\setlength{\tabcolsep}{8pt}
\renewcommand{\arraystretch}{1.2}
\resizebox{\columnwidth}{!}
{                
\begin{tabular}{lrrrr}
\toprule
Dataset & Replica & TUM-RGBD & ScanNet & ScanNet++ \\ \midrule
Translation $(\text{cm})$ & 1.07&	1.39&	1.34&	14.77 \\
Rotation $($$^{\circ})$ & 0.50&	1.37&	0.69&	13.43 \\
\bottomrule 
\end{tabular}
}
\caption{\textbf{Average Frame Motion Across Datasets}.}
\label{tab:datasets_avg_motion}
\end{table}

\section{Novel View Synthesis}
We evaluate the novel view synthesis (NVS) performance using the test set of the ScanNet++ sequences, where the test views are held-out and distant from training views. PSNR is evaluated on all test views after $10K$ iterations of global map refinement using the image resolution of $876\times584$. \cref{tab:eval_nvs} shows that ours yields the best NVS results. For the baselines, we implement the evaluation using their open-sourced code.
\begin{table}[htb]
\centering
\scriptsize
\renewcommand{\arraystretch}{1.0}

\begin{tabularx}{\columnwidth}{lRRRRRR}
\toprule
Method & \texttt{a} & \texttt{b} & \texttt{c} & \texttt{d} & \texttt{e} & Avg. \\
\midrule
ESLAM~\cite{mahdi2022eslam}& 13.63	&11.86	&11.83	&\rd10.59	&10.64	&11.71 \\
SplaTAM~\cite{keetha2023splatam} & \rd23.95 & \rd22.66 & \rd13.95 & 8.47 & \rd 20.06 & \rd 17.82 \\
Gaussian-SLAM~\cite{yugay2023gaussian} & \fs26.66 & \fs24.42	& \nd15.01	& \nd18.35	& \nd21.91	& \nd21.27 \\
\ours (Ours) & \nd25.60 &	\nd23.65 & \fs15.87&	\fs18.86 &	\fs22.51 &	\fs21.30 \\

\bottomrule
\end{tabularx}
\caption{\textbf{Novel View Synthesis on ScanNet++~\cite{yeshwanthliu2023scannetpp}} (PSNR $\uparrow$ [dB]). For the baselines, results were obtained using the open-sourced code with our implementation for the NVS evaluation. PSNR calculations include all pixels, regardless of whether they have valid depth input. \ours yields the best results.}

\label{tab:eval_nvs}
\end{table}

\section{Additional Analysis}

\begin{table}[!htb]
\centering
\scriptsize
\setlength{\tabcolsep}{2pt}
\renewcommand{\arraystretch}{1.5} %
\resizebox{\columnwidth}{!}{ 
\begin{tabular}{llccccccccc}
\toprule
Method & Metric & \texttt{Rm0} & \texttt{Rm1} & \texttt{Rm2} & \texttt{Off0} & \texttt{Off1} & \texttt{Off2} & \texttt{Off3} & \texttt{Off4} & Avg.\\
\midrule
\multirow{3}{*}{\makecell[l]{NICE-SLAM \cite{zhu2022nice}}}
& PSNR$\uparrow$ & 22.12 &  22.47 &  24.52 &  29.07 &  30.34 &  19.66 &  22.23 &  24.94 &  24.42 \\
& SSIM $\uparrow$ &  0.689 &   0.757 &  0.814 &  0.874 &   0.886 &  0.797 &  0.801 &  0.856 &  0.809 \\
& LPIPS $\downarrow$ &  0.330 &  0.271 &   0.208 &  0.229 &   0.181 &   0.235 &  0.209 &   0.198 &   0.233\\
[0.8pt] \hdashline \noalign{\vskip 1pt}
\multirow{3}{*}{\makecell[l]{Vox-Fusion \cite{yang2022vox}}} & PSNR$\uparrow$ &  22.39 &  22.36 &  23.92 &  27.79 &   29.83 &  20.33 &  23.47 &  25.21 &  24.41 \\
& SSIM$\uparrow$ &  0.683 &  0.751 &  0.798 &  0.857 &  0.876 &  0.794 &  0.803 &  0.847 &  0.801\\
& LPIPS$\downarrow$ &   0.303 &   0.269 &  0.234 &  0.241 &  0.184 &  0.243 &  0.213 &  0.199 &  0.236\\
[0.8pt] \hdashline \noalign{\vskip 1pt}
\multirow{3}{*}{\makecell[l]{ESLAM \cite{eslam_cvpr23}}} & PSNR$\uparrow$ &  25.25 &   27.39 &   28.09 &   30.33 &  27.04 &  27.99 &   29.27 &   29.15 &   28.06 \\
& SSIM$\uparrow$ &  0.874 &  0.89 &   0.935 &   0.934 &   0.910 &  0.942 &   0.953 &   0.948 &   0.923\\
& LPIPS$\downarrow$ & 0.315 &  0.296 &  0.245 &   0.213 &  0.254 & 0.238 &   0.186 &  0.210 & 0.245 \\
[0.8pt] \hdashline \noalign{\vskip 1pt}
\multirow{3}{*}{\makecell[l]{Point-SLAM \cite{sandstrom2023point}}}
& PSNR$\uparrow$  &  32.40 & 34.08 & 35.50  &  38.26 & 39.16 &  33.99 &  33.48 &  33.49 &  35.17\\
& SSIM$\uparrow$ &  0.974 &  0.977 &  0.982 &  0.983 &	 0.986 &   0.960 &   0.960 &   0.979 &   0.975 \\
& LPIPS$\downarrow$ &  0.113 &  0.116 &  0.111 &  0.100 &	  0.118 &  0.156 &  0.132 &  0.142 &   0.124 \\
[0.8pt] \hdashline \noalign{\vskip 1pt}
\multirow{3}{*}{\makecell[l]{SplaTAM~\cite{keetha2023splatam}}} 
& PSNR$\uparrow$  & 32.86 & 33.89 & 35.25 & 38.26 & 39.17 & 31.97 & 29.70 & 31.81 & 34.11 \\
& SSIM$\uparrow$ & 0.98 & 0.97 & 0.98 & 0.98 & 0.98 & 0.97 & 0.95 & 0.95 & 0.97 \\
& LPIPS$\downarrow$ & 0.07 & 0.10 & 0.08 & 0.09 & 0.09 & 0.10 & 0.12 & 0.15 & 0.10 \\
\hdashline \noalign{\vskip 1pt}
\multirow{3}{*}{\makecell[l]{\textcolor{red}{$^*$}Gaussian-SLAM~\cite{yugay2023gaussian}}} 
& PSNR$\uparrow$  & 38.88 & 41.80 & 42.44 & 46.40 & 45.29 & 40.10 & 39.06 & 42.65 & 42.08 \\
& SSIM$\uparrow$ & 0.993 & 0.996 & 0.996 & 0.998 & 0.997 & 0.997 & 0.997 & 0.997 & 0.996 \\
& LPIPS$\downarrow$ & 0.017 & 0.018 & 0.019 & 0.015 & 0.016 & 0.020 & 0.020 & 0.020 & 0.018 \\

\hdashline \noalign{\vskip 1pt}
\multirow{3}{*}{\makecell[l]{\ours} }
& PSNR$\uparrow$  & \fs33.07 & \fs 35.32&	\fs 36.16 &\fs40.82&	\fs40.21	&\fs34.67	&\fs35.67	& \fs37.10 & \fs 36.63 \\
& SSIM$\uparrow$ & \rd0.973&	\fs0.978&	\fs0.985	&\fs0.992	&\fs0.990	&\fs0.985	&\fs0.990	&\fs0.989&	\fs0.985 \\
& LPIPS$\downarrow$ & \rd0.116 &	\rd0.122	& \nd0.111&	\fs0.085	& \rd0.123&	\nd0.140&	\fs0.096&	\fs0.106&	\nd0.112 \\

\bottomrule
\end{tabular}}
\caption{\textbf{Rendering Performance on Replica~\cite{replica19arxiv}.} \textcolor{red}{$^*$} denotes evaluating on submaps instead of a global one.}
\label{tab:replica_rendering}
\end{table}

\begin{table}[!htb]
\centering
\resizebox{\columnwidth}{!}
{
\begin{tabular}{lcccccc}
\toprule
Method & Metric & \texttt{fr1/desk} & \texttt{fr2/xyz} & \texttt{fr3/office} & Avg.\\
\midrule
\multirow{3}{*}{\makecell[l]{NICE-SLAM~\cite{zhu2022nice}}}
& PSNR$\uparrow$ & 13.83 & 17.87 & 12.890 & 14.86 \\
& SSIM$\uparrow$ & 0.569 & 0.718 & 0.554 & 0.614 \\
& LPIPS$\downarrow$ & 0.482 & 0.344 & 0.498 & 0.441 \\[0.8pt] \hdashline \noalign{\vskip 1pt}
\multirow{3}{*}{\makecell[l]{Vox-Fusion~\cite{yang2022vox}}}
& PSNR$\uparrow$ & 15.79 & 16.32 & 17.27 & 16.46 \\
& SSIM$\uparrow$ & 0.647 & 0.706 & 0.677 & 0.677 \\
& LPIPS$\downarrow$ & 0.523 & 0.433 & 0.456 & 0.471 \\[0.8pt] \hdashline \noalign{\vskip 1pt}
\multirow{3}{*}{\makecell[l]{ESLAM~\cite{eslam_cvpr23}}}
& PSNR$\uparrow$ & 11.29 & 17.46 & 17.02 & 15.26 \\
& SSIM$\uparrow$ & 0.666 & 0.310 & 0.457 & 0.478 \\
& LPIPS$\downarrow$ & 0.358 & 0.698 & 0.652 & 0.569 \\[0.8pt] \hdashline \noalign{\vskip 1pt}
\multirow{3}{*}{\makecell[l]{Point-SLAM~\cite{sandstrom2023point}}}
& PSNR$\uparrow$ & 13.87 & 17.56 & 18.43 & 16.62 \\
& SSIM$\uparrow$ & 0.627 & 0.708 & 0.754 & 0.696 \\
& LPIPS$\downarrow$ & 0.544 & 0.585 & 0.448 & 0.526 \\[0.8pt] \hdashline \noalign{\vskip 1pt}
\multirow{3}{*}{\makecell[l]{SplaTAM~\cite{keetha2023splatam}}}
& PSNR$\uparrow$ & 22.00 & 24.50 & 21.90 & 22.80 \\
& SSIM$\uparrow$ & 0.857 & 0.947 & 0.876 & 0.893 \\
& LPIPS$\downarrow$ & 0.232 & 0.100 & 0.202 & 0.178 \\[0.8pt] \hdashline \noalign{\vskip 1pt}
\multirow{3}{*}{\makecell[l]{\textcolor{red}{$^*$}Gaussian-SLAM~\cite{yugay2023gaussian}}}
& PSNR$\uparrow$ & 24.01 & 25.02 & 26.13 & 25.05 \\
& SSIM$\uparrow$ & 0.924 & 0.924 & 0.939 & 0.929 \\
& LPIPS$\downarrow$ & 0.178 & 0.186 & 0.141 & 0.168 \\[0.8pt] \hdashline \noalign{\vskip 1pt}

\multirow{3}{*}{\makecell[l]{\ours}}
& PSNR$\uparrow$ &\fs22.03	& \nd22.68	& \fs23.47	&\nd22.72 \\
& SSIM$\uparrow$ &  \nd0.849	&\nd0.892&	\fs0.879&	\nd0.873\\
& LPIPS$\downarrow$ & \nd0.307	&\nd0.217	&\nd0.253&	\nd0.259 \\
\bottomrule
\end{tabular}
}
\caption{\textbf{Rendering Performance on TUM RGB-D~\cite{sturm2012benchmark}.} \textcolor{red}{$^*$} denotes evaluating on submaps instead of a global one.}
\label{tab:tum_rendering}
\end{table}

\begin{table}[!htb]
\centering
\setlength{\tabcolsep}{3pt}
\renewcommand{\arraystretch}{1.5} %
\resizebox{\columnwidth}{!}
{
\begin{tabular}{llccccccccc}
\toprule
Method & Metric & \texttt{0000} & \texttt{0059} & \texttt{0106} & \texttt{0169} & \texttt{0181} & \texttt{0207} & Avg.\\
\midrule
\multirow{3}{*}{\makecell[l]{NICE-SLAM~\cite{zhu2022nice}}}
& PSNR$\uparrow$ & 18.71 &  16.55 &  17.29 &  {18.75} & 15.56 & 18.38 & 17.54 \\
& SSIM$\uparrow$ & 0.641 & 0.605 & 0.646 & 0.629 & 0.562 & 0.646 & 0.621 \\
& LPIPS$\downarrow$ & 0.561 & 0.534 &  0.510 & 0.534 & 0.602 & 0.552 & 0.548 \\
\hdashline \noalign{\vskip 1pt}
\multirow{3}{*}{\makecell[l]{Vox-Fusion~\cite{yang2022vox}}} 
& PSNR$\uparrow$ & 19.06 & 16.38 &  {18.46} &  18.69 &  16.75 &  19.66 &  18.17 \\
& SSIM$\uparrow$ & 0.662 & 0.615 &  {0.753} & 0.650 & 0.666 &  {0.696} &  {0.673} \\
& LPIPS$\downarrow$ & 0.515 & 0.528 &  {0.439} &  0.513 & 0.532 &  {0.500} &  0.504 \\
\hdashline \noalign{\vskip 1pt}
\multirow{3}{*}{ESLAM~\cite{eslam_cvpr23}} 
& PSNR$\uparrow$ & 15.70 & 14.48 & 15.44 & 14.56 & 14.22 & 17.32 & 15.29 \\
& SSIM$\uparrow$ &  {0.687} &  0.632 & 0.628 &  0.656 &  {0.696} & 0.653 & 0.658 \\
& LPIPS$\downarrow$ &  {0.449} &  {0.450} & 0.529 &  {0.486} &  0.482 & 0.534 &  {0.488} \\
\hdashline \noalign{\vskip 1pt}
\multirow{3}{*}{\makecell[l]{Point-SLAM~\cite{sandstrom2023point}}} 
& PSNR$\uparrow$ &  {21.30} &  {19.48} & 16.80 & 18.53 &  {22.27} &  {20.56} &  19.82 \\
& SSIM$\uparrow$ &  {0.806} &  {0.765} &  {0.676} &  {0.686} &  {0.823} &  0.750 &  0.751 \\
& LPIPS$\downarrow$ &  0.485 &  0.499 & 0.544 & 0.542 &  {0.471} & 0.544 & 0.514 \\
\hdashline \noalign{\vskip 1pt}
\multirow{3}{*}{SplaTAM~\cite{keetha2023splatam}} 
& PSNR$\uparrow$ & {19.33} & {19.27} & {17.73} & {21.97} & 16.76 & {19.8} & 19.14 \\
& SSIM$\uparrow$ & 0.660 &  {0.792} & 0.690 &  {0.776} & 0.683 & 0.696 & 0.716 \\
& LPIPS$\downarrow$ & 0.438 &  {0.289} & {0.376} & {0.281} & {0.420} & {0.341} & 0.358 \\
\hdashline \noalign{\vskip 1pt}
\multirow{3}{*}{\textcolor{red}{$^*$}Gaussian-SLAM~\cite{yugay2023gaussian}} 
& PSNR$\uparrow$ & {28.539} & {26.208} & {26.258} & {28.604} & {27.789} & {28.627} & 27.67 \\
& SSIM$\uparrow$ & {0.926} & {0.9336} & {0.9259} & {0.917} & {0.9223} & {0.9135} & 0.923 \\
& LPIPS$\downarrow$ & {0.271} & {0.211} & {0.217} & {0.226} & {0.277} & {0.288} & 0.248 \\

\hdashline \noalign{\vskip 1pt}
\multirow{3}{*}{\ours (Ours)} 
& PSNR$\uparrow$ &  \fs24.99 &	\fs23.23&	\fs23.35 &\fs26.80&\fs24.82	&\fs26.33 &\fs24.92\\
& SSIM$\uparrow$ &  \fs0.840&	\fs0.831&	\fs0.846&	\fs0.877&	\fs0.824&	\fs0.854	&\fs0.845\\
& LPIPS$\downarrow$ & \nd0.450	&\nd0.400	&\nd0.409	&\nd0.346	&\nd0.514	&\nd0.430 &\nd0.425\\
\bottomrule
\end{tabular}}

\caption{\textbf{Rendering Performance on ScanNet~\cite{dai2017scannet}.} \textcolor{red}{$^*$} denotes evaluating on submaps instead of a global one. We exclude these results from the comparison for not being fair and for evaluating an easier setting.}
\label{tab:scannet_rendering}
\end{table}

\boldparagraph{Rendering Performance at Scene Level.} In the main paper, we only report the average rendering performance on each dataset. \cref{tab:replica_rendering}, \cref{tab:tum_rendering}, and \cref{tab:scannet_rendering} report the per-scene rendering performance on Replica, TUM RGB-D, and ScanNet, respectively. On Replica and ScanNet, \ours has the best performance on most of the scenes and on TUM RGB-D, \ours is only second to SplaTAM~\cite{keetha2023splatam}. 

\boldparagraph{Online LC.} 
\begin{table}[t]
\centering
\setlength{\tabcolsep}{12pt}
\renewcommand{\arraystretch}{1.1}
\resizebox{\columnwidth}{!}{
\begin{tabular}{cccc}
\toprule

LC Mode& Replica & ScanNet & TUM RGB-D\\
\midrule
Offline & 0.26& 15.27 & 12.54	 \\
Online & 0.26 & 8.39 & 3.33 \\

\bottomrule
\end{tabular}
}
\caption{\textbf{Ablation Study on Offline LC.} (ATE [cm]$\downarrow$)}

\label{tab:offline_lc}
\end{table}
We investigate the significance of applying LC and PGO online in \ours, as opposed to applying them only after the entire run concludes. The online mode, as presented in our main paper, continuously performs LC and PGO during the SLAM process. In contrast, the offline mode delays these operations until the input stream ends, applying them only once.
Results in \cref{tab:offline_lc} reveal that for smaller scenes, such as those in Replica, online LC does not significantly improve performance due to the limited number of loops. However, in more complex environments like ScanNet and TUM RGB-D, online LC proves crucial to \ours's superior performance. This is because it constantly corrects map drift, preventing cumulative errors that would otherwise degrade accuracy over time.

\boldparagraph{Average Number of Gaussians Per Scene.} \cref{tab:num_3dgs} reports the average number of Gaussians after global map refinement for each dataset. For a room-sized scene, we obtain on average around $300K$ Gaussian splats, which is a reasonable number. The number of Gaussians is dependent on the scale of the scenes, the number of vertices used to initialize the Gaussians, and the number of densification iterations during the optimization of 3DGS.

\begin{table}[t]
\centering
\setlength{\tabcolsep}{8pt}
\renewcommand{\arraystretch}{1.2}
\resizebox{\columnwidth}{!}
{                
\begin{tabular}{lrrrr}
\toprule
Dataset & Replica & TUM-RGBD & ScanNet & ScanNet++ \\ \midrule
$\#$ Gaussians & $295K$ & $219K$	&	$331K$ & $330K$\\
\bottomrule 
\end{tabular}
}
\caption{\textbf{Average Number of Gaussians Per-scene}.}
\label{tab:num_3dgs}
\end{table}

\section{Additional Qualitative Results} 
\begin{figure*}[t]
    \centering

    \begin{tabular}{ccc}
    \multicolumn{3}{c}{%
        \includegraphics[width=\linewidth]{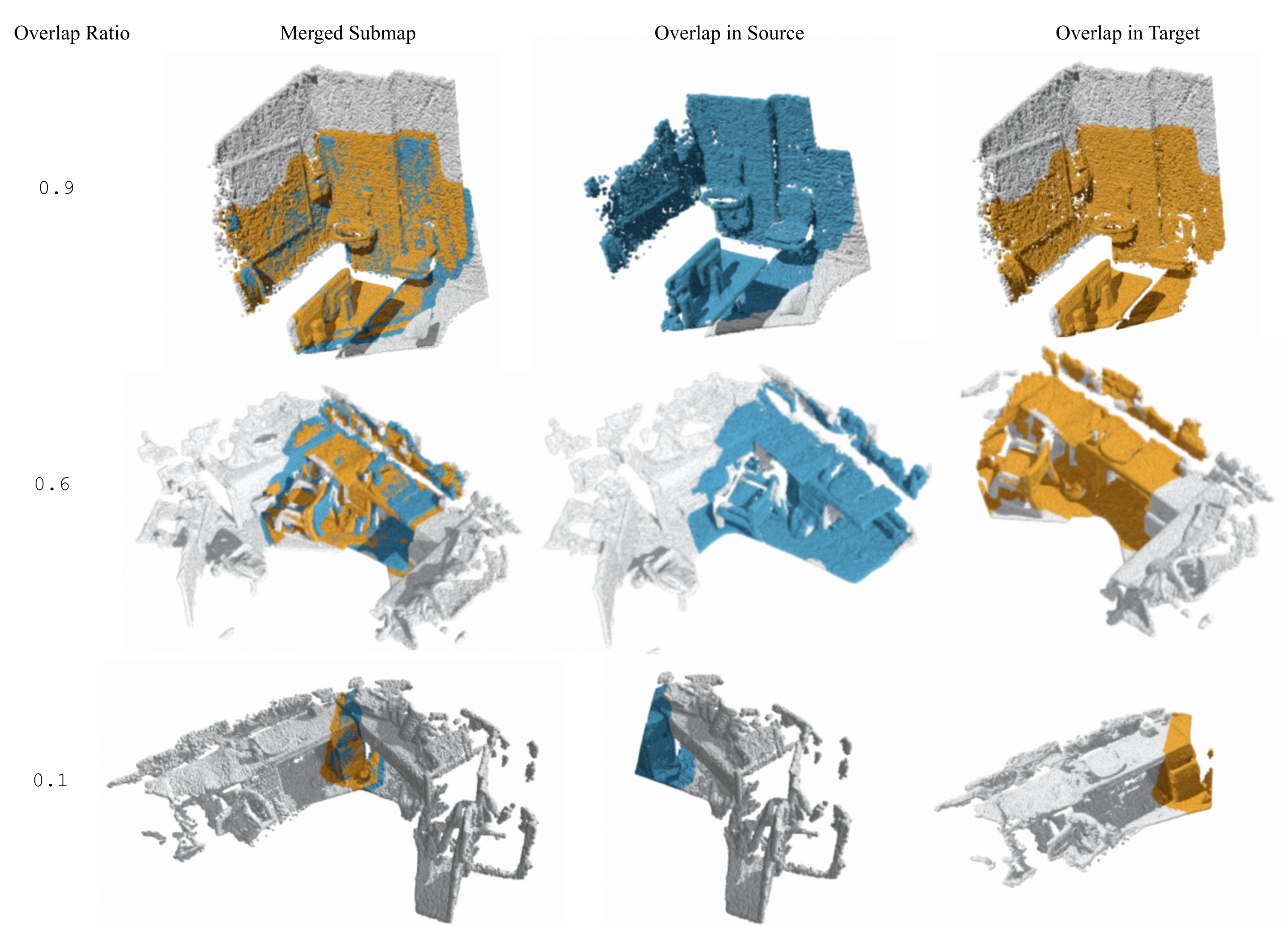}%
    } \\

\end{tabular}
\caption{\textbf{Qualitative Results of Overlap Ratio between Submaps}. We visualize the centers of 3D Gaussians as point clouds, with two submaps only colorized in the overlapping region. The top row demonstrates a large overlap between submaps with $\mathrm{OR}=0.9$. The middle row showcases a medium overlap of $\mathrm{OR}=0.6$, while the bottom row exhibits an extremely low overlap of $\mathrm{OR}=0.1$. This last case was rejected as a loop due to its insufficient overlap, which typically leads to low-accuracy registration or even complete failure.}
\label{fig:supp_overlap}
\end{figure*}

\begin{figure*}[t]
    \centering

    \begin{tabular}{ccc}
    \multicolumn{3}{c}{%
        \includegraphics[width=\linewidth]{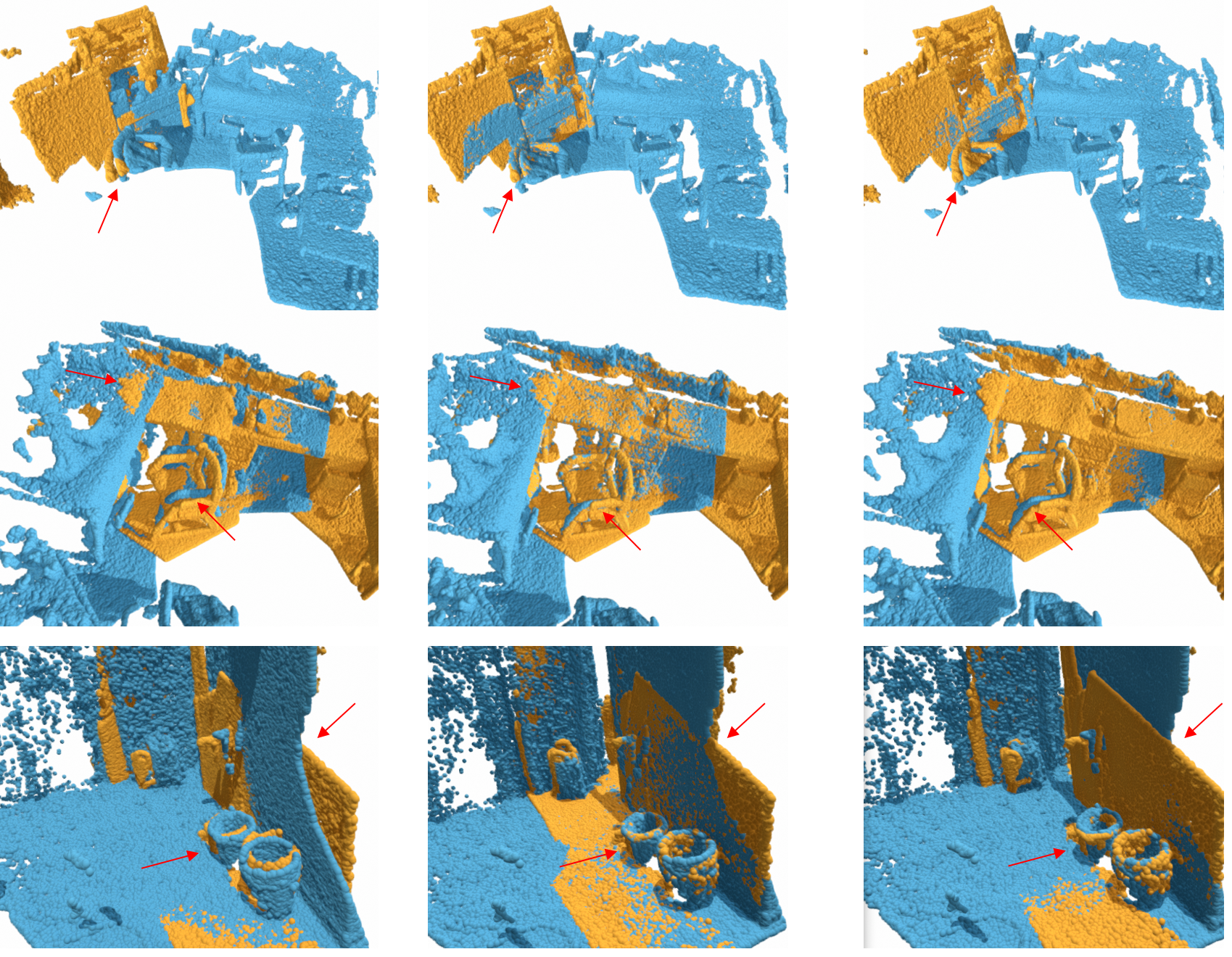}%
    } \\
    \small{\hspace{2.cm}Odometry} & \small{\hspace{3.8cm} \ours (Ours)} & \small{\hspace{2.5cm}Ground Truth} \\
\end{tabular}

    \caption{\textbf{Qualitative Results on Submap Registration}. We visualize the centers of 3D Gaussians as point clouds, with two submaps colorized differently. \ours consistently improves upon the initial odometry-based alignment and outperforms the pseudo ground truth. In the first row, \ours (middle) achieves better alignment of the chair's back compared to both odometry and ground truth. Similar improvements are observed in the second row. The last row demonstrates \ours's superior alignment of walls and trash cans. These results, representative of ScanNet and not cherry-picked, consistently showcase the method's effectiveness across various scenes.
    }
    \label{fig:supp_reg}
\end{figure*}

\begin{figure*}[t]
    \centering
 \begin{tabular}{ccccc}
    \multicolumn{5}{c}{%
        \includegraphics[width=\linewidth]{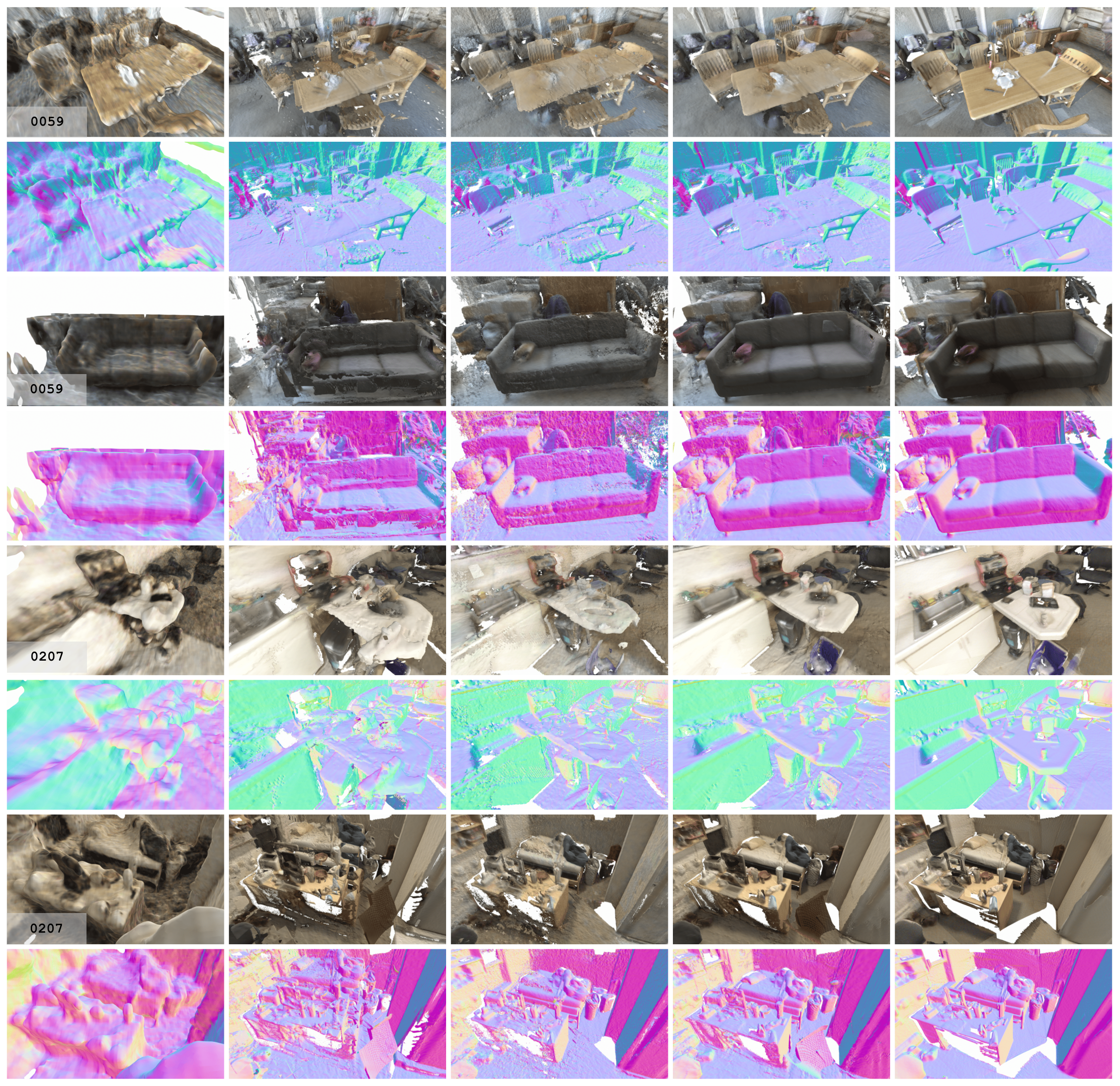}%
    } \\
    \hspace{1cm} \small GO-SLAM~\cite{zhang2023goslam} & \hspace{0.45cm} \small Gaussian-SLAM~\cite{yugay2023gaussian} & \hspace{0.4cm}\small Loopy-SLAM~\cite{liso2024loopyslam} & \hspace{0.7cm}\small  LoopSplat (Ours) & \small Ground Truth \\
\end{tabular}  
    
    \caption{\textbf{Mesh Reconstruction on ScanNet~\cite{dai2017scannet} scenes \texttt{0059} and \texttt{0207}}. Per example, the first row displays the colored mesh, while the second row shows the corresponding normals. \ours demonstrates superior performance compared to baseline methods, excelling in both texture fidelity and geometric detail. Notably, our approach yields smoother and more complete mesh reconstructions than the strongest baseline, Loopy-SLAM.}
    
    \label{fig:recon}
\label{fig:supp_recon}

\end{figure*}

In this section, we present additional qualitative results. 
\paragraph{Overlap Ratio.} We first illustrate the overlap ratio we adopt to determine if a detected loop is added to the pose graph. In \cref{fig:supp_overlap}, we showcase three representative ScanNet submap pairs with descending overlap ratios.
\paragraph{3DGS Registration.} \cref{fig:supp_reg} presents more registration results on the submaps. The red arrows highlight the differences between odometry, ours, and ground truth. The odometry results have the most misalignment, whereas estimates from \ours are closer to, or even better than, the ground truth through visual inspection.
\paragraph{Mesh Reconstruction.} We present additional qualitative results for mesh reconstruction on ScanNet scenes \texttt{0059} and \texttt{0207} in \cref{fig:supp_recon}. Our analysis concentrates on regions with high geometric complexity. As evident from the results, \ours consistently produces higher-quality and more consistent reconstructions compared to baseline methods, particularly in these challenging areas.

\section{Limitations and Future Work}
\paragraph{Limitations.} \ours still faces certain limitations. As the number of submaps exceeds 100, the computational demands for pairwise registrations during pose graph optimization increase significantly, reducing the efficiency of the loop closure module. While \ours demonstrates competitive performance and achieves the lowest peak GPU usage among all compared methods, there remains significant room to improve the system's overall efficiency. The iterative nature of optimizing 3D Gaussians and camera poses limits the speed of the system. The pose initialization is based on the constant speed assumption, which can be improved with Kalman Filters. In terms of submap construction, we use different hyperparameters for different datasets, which is a standard practice in the SLAM community, but we believe it hinders the generalization ability of the system to in-the-wild data.

\paragraph{Future Work.} Several promising avenues for future research emerge from this work. First, employing advanced mesh extraction methods that directly operate on 3DGS, such as SuGAR~\cite{guedon2023sugar} or GOF~\cite{Yu2024GOF}, can improve the reconstruction performance. Second, integrating uncertainty estimates for each viewpoint could improve both overlap estimation and multi-view optimization in 3DGS registration. Additionally, exploring techniques to refine 3DGS reconstruction in overlapping regions between submaps presents another intriguing direction. 

\end{document}